\documentclass[conference]{IEEEtran}
\IEEEoverridecommandlockouts
\usepackage{cite}
\usepackage{amsmath, amsthm, amssymb,amsfonts}
\usepackage{bm}
\usepackage{algorithmic}
\usepackage[ruled,linesnumbered]{algorithm2e}
\usepackage{graphicx}
\usepackage{textcomp}
\usepackage{subfigure}
\usepackage{booktabs}
\usepackage{colortbl}
\usepackage{soul}
\usepackage{xcolor}
\newtheorem{remark}{Remark}

\def\BibTeX{{\rm B\kern-.05em{\sc i\kern-.025em b}\kern-.08em
    T\kern-.1667em\lower.7ex\hbox{E}\kern-.125emX}}
\SetKwBlock{uTry}{try}{}
\SetKwBlock{uCatch}{catch}{end}
\begin{document}

\title{Optimization-Based Trajectory Planning for Tractor-Trailer Vehicles on Curvy Roads: A Progressively Increasing Sampling Number Method
\thanks{
All authors are with Department of Automation, Institute of Medical Robotics, Shanghai Jiao Tong University, and Key Laboratory of System Control and Information Processing, Ministry of Education of China, and Shanghai Engineering Research Center of Intelligent Control and Management, Shanghai 200240, China. Han Zhang is the corresponding author (email: zhanghan\_tc@sjtu.edu.cn).

This work was supported by the National Key R\&D Program of China under Grant 2022YFC3601403, and the ZF (China) Investment Co., Ltd.}
}

\author{
\IEEEauthorblockN{Zehao Wang, Han Zhang, Jingchuan Wang, and Weidong Chen}
}

\maketitle

\begin{abstract}
In this work, we propose an optimization-based trajectory planner for tractor-trailer vehicles on curvy roads. The lack of analytical expression for the trailer's errors to the center line pose a great challenge to the trajectory planning for tractor-trailer vehicles. To address this issue, we first use geometric representations to characterize the lateral and orientation errors in Cartesian frame, where the errors would serve as the components of the cost function and the road edge constraints within our optimization process. Next, we generate a coarse trajectory to warm-start the subsequent optimization problems. 
On the other hand, to achieve a good approximation of the continuous-time kinematics, optimization-based methods usually discretize the kinematics with a large sampling number. This leads to an increase in the number of the variables and constraints, thus making the optimization problem difficult to solve. To address this issue, we design a Progressively Increasing Sampling Number Optimization (PISNO) framework. More specifically, we first find a nearly feasible trajectory with a small sampling number to warm-start the optimization process. Then, the sampling number is progressively increased, and the corresponding intermediate Optimal Control Problem (OCP) is solved in each iteration. Next, we further resample the obtained solution into a finer sampling period, and then use it to warm-start the intermediate OCP in next iteration. This process is repeated until reaching a threshold sampling number. Simulation and experiment results show the proposed method exhibits a good performance and less computational consumption over the benchmarks.
\end{abstract}


\section{Introduction}
\subsection{Background}
In recent years, tractor-trailer vehicles are widely used in agriculture and logistics due to their large cargo capacity, high transport efficiency, and low fuel consumption due to their large cargo capacity, high transport efficiency and low fuel consumption \cite{ritzen2015,zhao2021,chai2024cooperative, wang2020review}. However, due to the vehicle's underactuated and counterintuitive kinematics \cite{li2015robio}, maneuvering a tractor-trailer vehicle is still a challenging task. It could be more challenging when the curvature of the road and the number of obstacles increases \cite{jujnovich2013}. To address this issue, we focus on the trajectory planning task for tractor-trailer vehicles on curvy roads in this work. 

\subsection{Related works}
Existing trajectory planners for vehicles are divided into three categories: sample-and-search-based, optimization-based and learning-based methods. Although several studies \cite{chai2022design,chai2022deep,chai2020design,zhang2024efficient} demonstrate the potential of Deep Neural Network (DNN) and Reinforcement Learning (RL) in solving the trajectory planning problems, however, training an effective planning policy through DNN and RL typically requires extensive offline training. Moreover, these methods require robust training algorithms to ensure the convergence. In addition, these methods typically lack interpretability, which is a critical concern in autonomous driving under the safety requirements.

Sample-and-search-based methods first sample the state or control space, then search for a feasible path within the space. Existing researches exploit the hybrid A* \cite{beyersdorfer2013}, Rapidly Exploring Random Tree (RRT) \cite{evestedt2016, manav2021, zhao2023} and motion primitives \cite{rimmer2016, ljungqvist2017, wang2023, leu2022} to find a feasible path. These methods have merits in computational efficiency over the optimization-based methods.  
However, the outcome of such planning methods is a path rather than a trajectory, thereby lacking the spatial-temporal information. Also, these methods cannot guarantee the optimality and smoothness of the trajectory, which are critical for the on-road planning. Moreover, the sampling process inevitably discards some potential feasible solutions. Therefore, in narrow scenarios, these methods require more sampling times, resulting in being less efficient and may even fail in providing a feasible path \cite{li2021}.

On the other hand, optimization-based methods typically formulate the task as an Optimal Control Problem (OCP), then convert the OCP into a Nonlinear Programming (NLP) problem and solve it numerically. Li et al. \cite{li2015} use the triangular-area-based collision-avoidance constraints to define the OCP and then employ an incremental initialization strategy \cite{li2015robio} to find a feasible solution. Bergman et al. \cite{bergman2020} implement a bilevel optimization framework to find a local optimal solution based on the motion primitives. Moreover, Li et al. \cite{li2019,li2019icra} design strategy to progressive add the collision-avoidance constraints. Cen et al. \cite{cen2021} implement linear spatio-temporal corridors to replace the large-scale triangle-area-based collision-avoidance constraints. Wang et al. \cite{liu2022homogenization, wang2024safe} also design linear safety dispatch constraints (SDC) to simplify the collision-avoidance constraints. However, the linear collision-avoidance constraints would discard some feasible regions, resulting in lower success rate. Li et al. \cite{li2021} first soften the kinematic constraints, and then use an iterative framework to refine the kinematic feasibility. 
Optimization-based methods have advantages in the trajectory quality, however, existing optimization-based methods try to directly solve a large-scale optimization problem. Although some technics are implemented to find a high quality initial guess for warm-start, however finding such kinematically feasible and collision-free trajectory as the initial guess is difficult and time-consuming. As a result, these methods often have drawbacks in computation costs.

Besides the unstructured scenarios, few studies have been focused on the tractor-trailer's trajectory planning methods for on-road scenarios. Existing on-road planners for passenger vehicles rely on the Frenet frame. However, the Frenet frame fails to model the vehicle's kinematics and ignores the distortion of the vehicle shape \cite{li2022}. The situation would get worse for tractor-trailer vehicles because the trailer's lateral and orientation errors cannot be expressed analytically \cite{oliveira2020}. To address this issue, \cite{oliveira2020, shen2021, oliveira2020iv} first approximate the errors, then solve the trajectory planning problem through nonlinear programming. However, the state estimation errors caused by the approximation would lead to non-optimality in the subsequent optimization. In addition, \cite{oliveira2020, shen2021, oliveira2020iv} do not exploit the computational advantage brought by the Frenet frame \cite{zhang2020}. In contrast, Cartesian frame is more suitable for tractor-trailer vehicles as it could accurately model the vehicle kinematics.

\subsection{Motivations and contributions}
As a summary, previous on-road trajectory planner for tractor-trailer vehicles fail to model the lateral and orientation errors accurately.
Moreover, existing optimization-based methods need to discretize the OCP with a considerable sampling number, thereby resulting in large computation costs. To address this issue, based on the hp-mesh refinement techniques \cite{peng2017hp}, we propose the Progressively Increasing Sampling Number Optimization (PISNO) framework. In particular, we use the following techniques to formulate the OCP and fasten the computation process:
\begin{itemize}
    \item We characterize the lateral and orientation errors for tractor-trailer vehicle in Cartesian coordinate.
    \item We use a new coarse trajectory generation method to determine the trajectory homotopy class, and giving a good initial guess for the subsequent optimization.
    \item We propose an iterative framework to solve the large-scale optimization problem by progressively increasing the sampling number for warm-start.
\end{itemize}

\section{Problem formulation}
When dealing with the on-road trajectory optimizations, a typical method is dividing the maneuver process into multiple phases \cite{chai2022multiphase}. Although this multiphase formulation could provide more feasibility for the optimization process, multiphase-based methods would introduce more additional parameters and decision variables to the optimization process. As the number of phases would increase, these methods would result in a higher computational cost. Thus, we formulate the on-road trajectory planning task for tractor-trailer vehicles as a single phase discretized OCP.

\subsection{The vehicle kinematics and the boundary constraint}
\begin{figure}[!htbp]
\centerline{\includegraphics[width=0.8\linewidth]{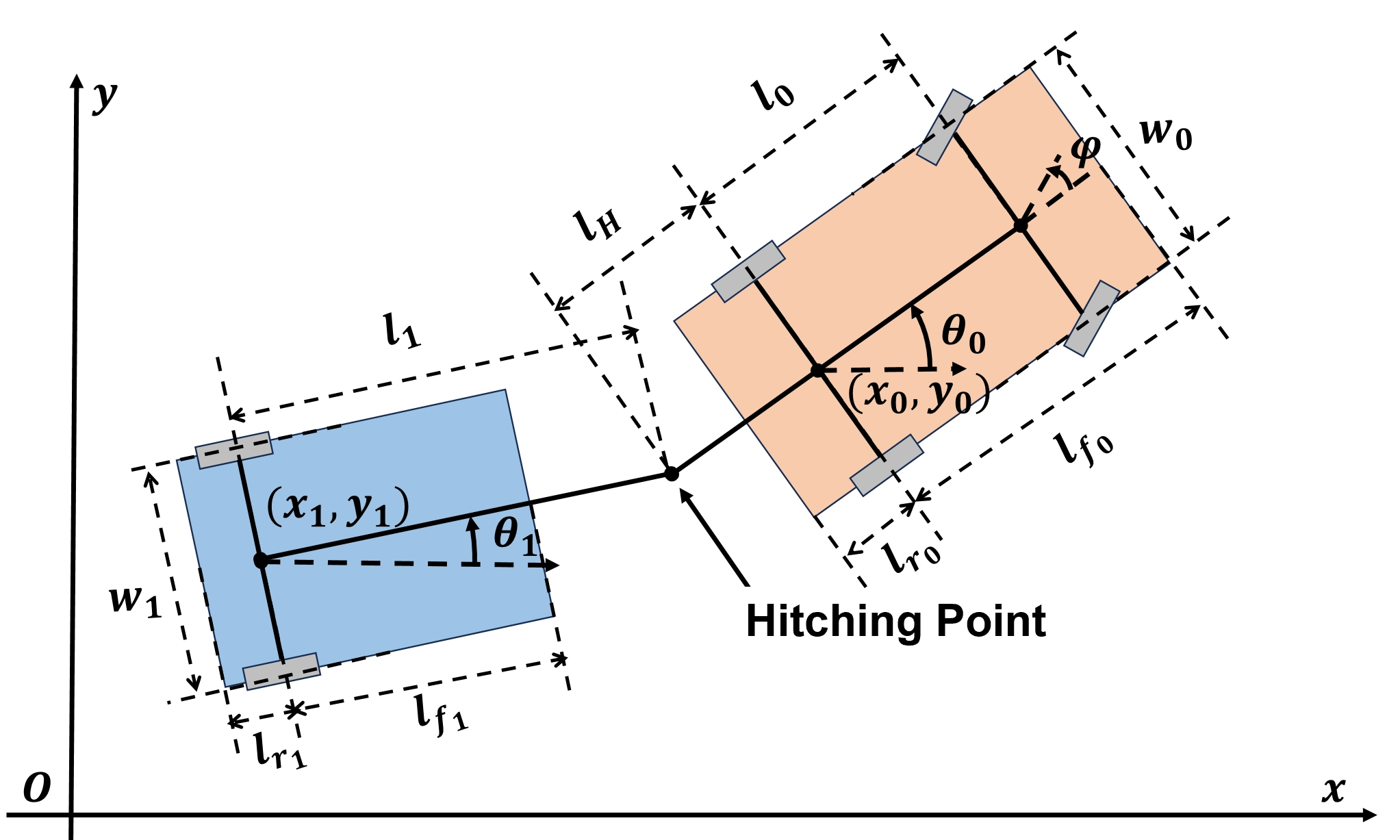}}
\caption{The schematic of a tractor-trailer vehicle system.}
\label{tractor trailer fig}
\end{figure}

As depicted in Fig. \ref{tractor trailer fig}, the tractor-trailer vehicle system consists a tractor towing a trailer with an off-axle hitch. 
The lengths of the tractor and trailer are denoted by $l_0$ and $l_1$ respectively, and $l_H$ represents the hitching offset distance. $l_{f_0}(l_{f_1})$, $l_{r_0}(l_{r_1})$ and $w_0(w_1)$ are the front overhang length, rear overhang length and  width of the tractor (trailer), respectively.
The trajectory of the tractor-trailer vehicle is discretized with a sampling period $\Delta t = T/N$, where $T$ denotes the planning time horizon and $N$ denotes the sampling number. 
Typically, $N$ is set to a relative large number, which is larger or equal to a threshold $N \geq N_{\rm thre}$ to well-approximate the kinematics in continuous-time. 
To ease the computational cost, we choose the bicycle model to describe the tractor's kinematics and discretize the vehicle kinematics as follows
\begin{equation}
\begin{aligned}
\begin{bmatrix}
\Delta x_0(k)\\
\Delta y_0(k)\\
\Delta \theta_0(k)\\
\Delta \theta_1(k)\\
\\
\Delta \varphi(k)\\
\Delta v(k)\\
\end{bmatrix}
&= \begin{bmatrix}
v(k) \cos\theta_0(k) \cdot \Delta t\\
v(k) \sin\theta_0(k) \cdot \Delta t \\
v(k) \tan \varphi(k)/ l_0 \cdot \Delta t \\
v(k) \sin(\theta_0(k)-\theta_1(k)) / l_1  \cdot \Delta t - \\
\Delta \theta_0(k) \cos(\theta_0(k)-\theta_1(k)) \cdot l_H / l_1 \\
\omega(k) \cdot \Delta t \\
a(k) \cdot \Delta t \\
\end{bmatrix}\!,\\
\ k=0,\dots, & \ N-1,\\
\end{aligned}
\label{kinematic model of tractor-trailer}
\end{equation}
where we use the parenthesis $(k)$ to denote the time instant in the discretized time series, $\Delta$ is the difference between the adjacent time-steps, namely, $\Delta x_0(k) = x_0(k+1) - x_0(k)$, for instance.
$(x_0, y_0)$ is the position of the tractor, $\theta_0$ and $\theta_1$ are the orientation angle of tractor and trailer respectively. $\varphi$ is the steering angle, $v$ is the linear velocity of the tractor's back wheels. $\omega$ is steering angle velocity and $a$ is the tractor's linear acceleration. In addition, the trailer's position
$(x_1, y_1)$ is computed as follows.
\begin{equation}
\label{trailer position}
\begin{aligned}
&x_1(k) \!=\! x_0(k) \!-\! l_H\!\cos\theta_0(k) \!-\! l_1\cos\theta_1(k),\\
&y_1(k) \!= y_0(k) \!-\! l_H\sin\theta_0(k) \!-\! l_1\sin\theta_1(k),k \!=\! 0,\dots, \!N.\\
\end{aligned}
\end{equation}
We further denote the full-state of the tractor-trailer vehicle system as $x(k) = [x_0(k), y_0(k), \theta_0(k), \theta_1(k), x_1(k), y_1(k), $ $\varphi(k), v(k)]^T$ and the control inputs are denoted as $u(k) = [\omega(k), a(k)]^T$.
Moreover, due to the physical constraints of the tractor-trailer vehicles, the planning problem is subjected to the following state and control bounds: 
\begin{equation}
\label{state and control bounds}
\begin{aligned}
-\delta\theta_{max} \leq & \theta_0(k) -\theta_1(k) \leq \delta\theta_{max}, \\
-\varphi_{max} \leq & \varphi(k) \leq \varphi_{max}, \\
-v_{max} \leq &v(k) \leq v_{max}, k = 0,\dots, N,\\
-\omega_{max} \leq &\omega(k) \leq \omega_{max}, \\
-a_{max} \leq &a(k) \leq a_{max}, k = 0,\dots, N-1,\\
\end{aligned}
\end{equation}
where $\varphi_{max}, v_{max},   \omega_{max}$ and $a_{max}$ represent the maximum permissible values for $\varphi, v, \omega$ and $a$, respectively. $\delta\theta_{max}$ is maximum jack-knife angle to guarantee the safety.

Moreover, we let the states at the start of the planning horizon be the tractor-trailer's current states $x^{\rm{start}}$. Namely, 
\begin{equation}
\label{Boundary Constraint}
x(0) \!=\! x^{\rm{start}}.
\end{equation}
In the on-road scenario, to keep more flexibility for the planner, we do not design any equality constraints for the terminal boundary state.

\subsection{The feasible region constraints}
As we focus on the on-road trajectory planing, convex polygons are used to model the obstacles such as static vehicles.  
The tractor-trailer vehicle needs to avoid the collision with each of the
obstacles at all time instants $ k \in \{0,\dots, N\}$. 
We denote the four vertexes of the tractor/trailer body at time instant $k$ as $\mbox{Veh}_i(k):=\{A_i(k), B_i(k), C_i(k), D_i(k)\},i\in\{0,1\}$, where $i=0$ and $i=1$ represents the body of the tractor and the trailer respectively.
The vertexes of $j$-th obstacle are denoted as $\mbox{Obs}_j:=\{V_j^1, \dots, V_j^{N_{j}}\}$, $j \in \{1,..., N_{obs}\}$, where  $N_{obs}$ is the number of obstacles and ${N_{j}}$ is the number of $j$-th obstacle's vertexes. 
Thus, for time-step $k$, the collision-avoidance constraints take the form
\begin{equation}
\label{Collision-Avoidance Constraints}
\begin{aligned}
&{\rm OutPoly}(p, \mbox{Obs}_j),  p\! \in\! \mbox{Veh}_i(k),\\
&{\rm OutPoly}(q, \mbox{Veh}_i(k)),   q\! \in\! \mbox{Obs}_j,\\
& k = 1, \ldots, N, \: i\!=\!0,1, \: j\!=\!1, \ldots, N_{obs},
\end{aligned}
\end{equation}
where ${\rm OutPoly}(a, \{b_1, \dots, b_n\})$ returns the logic value of whether the point $a$ is outside the convex polygon $\{b_1, \dots, b_n\}$. To this end, we choose the triangle-area criterion \cite{li2015} to describe ${\rm OutPoly}(a, \{b_1, \dots, b_n\})$. Moreover, the vehicle needs to avoid being outside the road edge:
\begin{equation}
\label{eq: road edge constraints}
\begin{aligned}
&\mbox{Veh}_i(k) \in \mathcal{F},\\
& k = 1, \ldots, N, \: i\!=\!0,1,
\end{aligned}
\end{equation}
where $\mathcal{F}$ is the region inside the road edges. The analytic expression of \eqref{eq: road edge constraints} will be detailed in \ref{sec:geometric_eq_and_errors}.

\subsection{The cost function}
In the on-road scenario, 
the cost function $J(\bm{x}, \bm{u})$ is composed of the cost to goal, lateral and orientation errors to the center line and the cost of control inputs, where the bolded vector $\bm{x} = [x(0)^T, x(1)^T, \dots, x(N)]^T$ and $\bm{u} = [u(0)^T, u(1)^T, \dots, u(N-1)]^T$ are used to represent the states and control inputs of all time-steps. A popular technique in constructing the cost function is to design multiple objectives and find the pareto solutions \cite{chai2020multiobjective, chai2020multiobjectiveoptimal,chai2024two}. However, this would introduce extra complexity to the optimization problem. Therefore, we choose to optimize a weighted objective function for brevity. In particular, it takes the form
\begin{equation}
\label{Cost Function}
\begin{aligned}
&J(\bm{x}, \bm{u})  = \omega_g ||x(N) - x^{g}||_W^2 \\
&+\sum_{k=0}^{N} \Delta t \cdot (\omega_{e} ||e(k)||^2) \!+ \!\omega_c ||u(k)||^2),  
\end{aligned}
\end{equation}
where $\omega_g, \omega_{e}, \omega_c$ are the trade-off weights.
More precisely, the term
$ ||x(N) - x^{g}||_W^2 := \sum_{i=0}^1(x_i(N) - x_i^{g})^2 + (y_i(N) - y_i^{g})^2 +  \omega_\theta(\theta_i(N) - \theta_i^{g})^2 $ is used to encourage the vehicle to be close to the goal $x^{g} := [x_0^{g}, y_0^{g}, \theta_0^{g}, x_1^{g}, y_1^{g}, \theta_1^{g}]^T$, where $\omega_\theta$ is the trade-off weight of orientation.
Moreover, in an on-road scenario, the tractor-trailer needs to keep close to the center line while avoiding the obstacles. Thus, the term $||e(k)||^2 := \sum_{i=0}^1 e_{p_i}(k)^2 +  \omega_\theta e_{\theta_i}(k)^2 $ is used to characterize the errors to the center line, where $e_{p_i}(k)$ and $e_{\theta_i}(k)$ represent the lateral and orientation errors of the tractor and the trailer to the center line at $k$-th time-step, respectively.  
In addition, $||u(k)||^2 = a(k)^2 + \omega(k)^2 $ represents the control cost, which guarantees the smoothness of the planned trajectory and saves energy.

\subsection{The overall OCP formulation}
\label{section 2e}
In summary, the on-road trajectory planning task for tractor-trailer vehicles is described by the following OCP
\begin{equation}
\label{Overall Formulation}
\tag{OCP}
\begin{aligned}
\min_{\substack{\bm{x},\bm{u}}}\;\;\;\; & \mbox{Cost function } \eqref{Cost Function},\\ 
\mbox{s.t.\;\;\;\;\;}& \rm Kinematic \ constraints \ \eqref{kinematic model of tractor-trailer}, \eqref{trailer position}, \eqref{state and control bounds}, \\
&\rm Boundary \ constraint \  \eqref{Boundary Constraint}, \\
&\rm Collision \ avoidance \ constraints \ \eqref{Collision-Avoidance Constraints},\\
&\rm Road \ edge \ constraints \ \eqref{eq: road edge constraints}.\\
\end{aligned}
\end{equation}
Note that, it is time-consuming and difficult to directly solve \eqref{Overall Formulation}. This is mainly because: (i) There is no explicit formulation to compute the lateral and orientation errors  \cite{oliveira2020}; (ii) We do not have a good initial guess for the numerical optimization solver. To address these difficulties, we design our trajectory planner in Section \ref{section 3}. 
\section{Trajectory planning method}
\label{section 3}

\subsection{Characterizing the lateral and orientation errors}
\label{sec:geometric_eq_and_errors}
Before continue, we need to characterize the lateral errors $e_{p_i}(k)$ and orientation errors $e_{\theta_i}(k)$ in the cost function \eqref{Cost Function}, and then give explicit expressions for the road edge constraints \eqref{eq: road edge constraints}. First, we assume the center line is composed of two key parts: straight and arc segments. In practice, the center line is discretized spatially with length resolution $\delta s$ as $\{(x^{\rm cen}[k], y^{\rm cen}[k], \theta^{\rm cen}[k])\}_{k=0}^{N_{\mathcal{C}}}=:\mathcal{C}$, where we use the bracket $[k]$ to denote the index in the spatially discretized series, $N_{\mathcal{C}}=:|\mathcal{C}|$ is the number of the points in the set.

Recall that the positions of the tractor-trailer vehicle $\{(x_i(k),y_i(k)), i \in \{0,1\}\}_{k=0}^{N}$ are the variables to be solved in the \eqref{Overall Formulation}, thus we do not know which center line segment that the vehicle's position project on. 
To address this issue, we use the initial guess $\{(x_{i}^{\rm init}(k), y_{i}^{\rm init}(k)), i \in \{0,1\}\}_{k=0}^{N}$ of the tractor-trailer positions 
where we assume that the projection of the initial guess would fall on the same center line segment as $\{(x_i(k),y_i(k)), i \in \{0,1\}\}_{k=0}^{N}$ would. 
We further denote the projection of the initial guess on the center line as $\{(x^{\rm cen}[n_{i,k}], y^{\rm cen}[n_{i,k}], \theta^{\rm cen}[n_{i,k}]), i \in \{0,1\}\}_{k=0}^{N} =: \mathcal{R}$, where $n_{i,k}$ is the index of the center point which the tractor/trailer initial guess position $(x_{i}^{\rm init}(k), y_{i}^{\rm init}(k))$ project onto.
\begin{figure}[!htbp]
\centering
\subfigure[The case when the center line is a straight line. We illustrate the tractor as an example.]{\includegraphics[width=1\hsize]{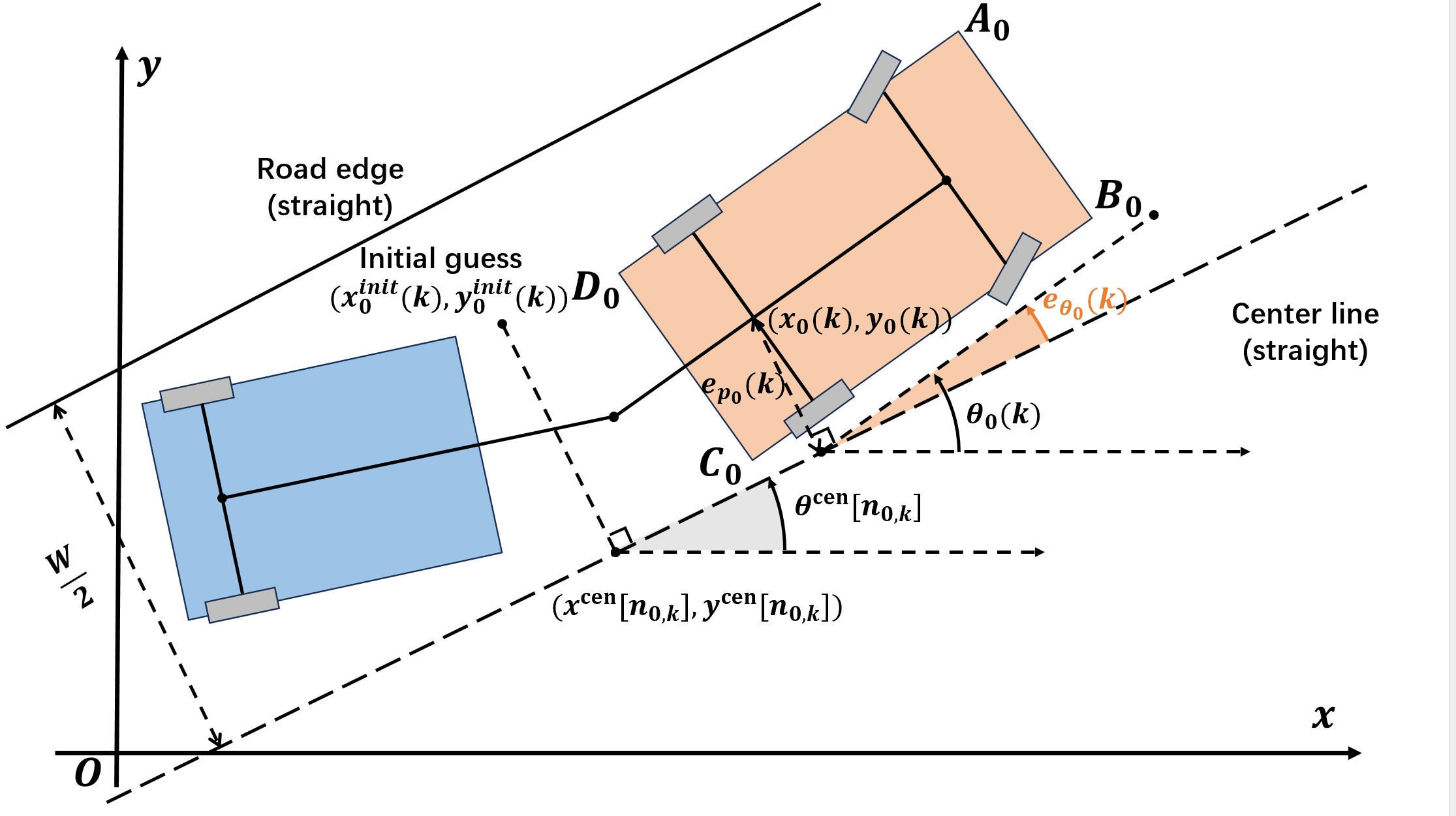}
\label{Center line is straight line. Take the tractor as an example}}
\subfigure[The case when the center line is an arc. We illustrate the trailer as an example.]
{\includegraphics[width=1\hsize]{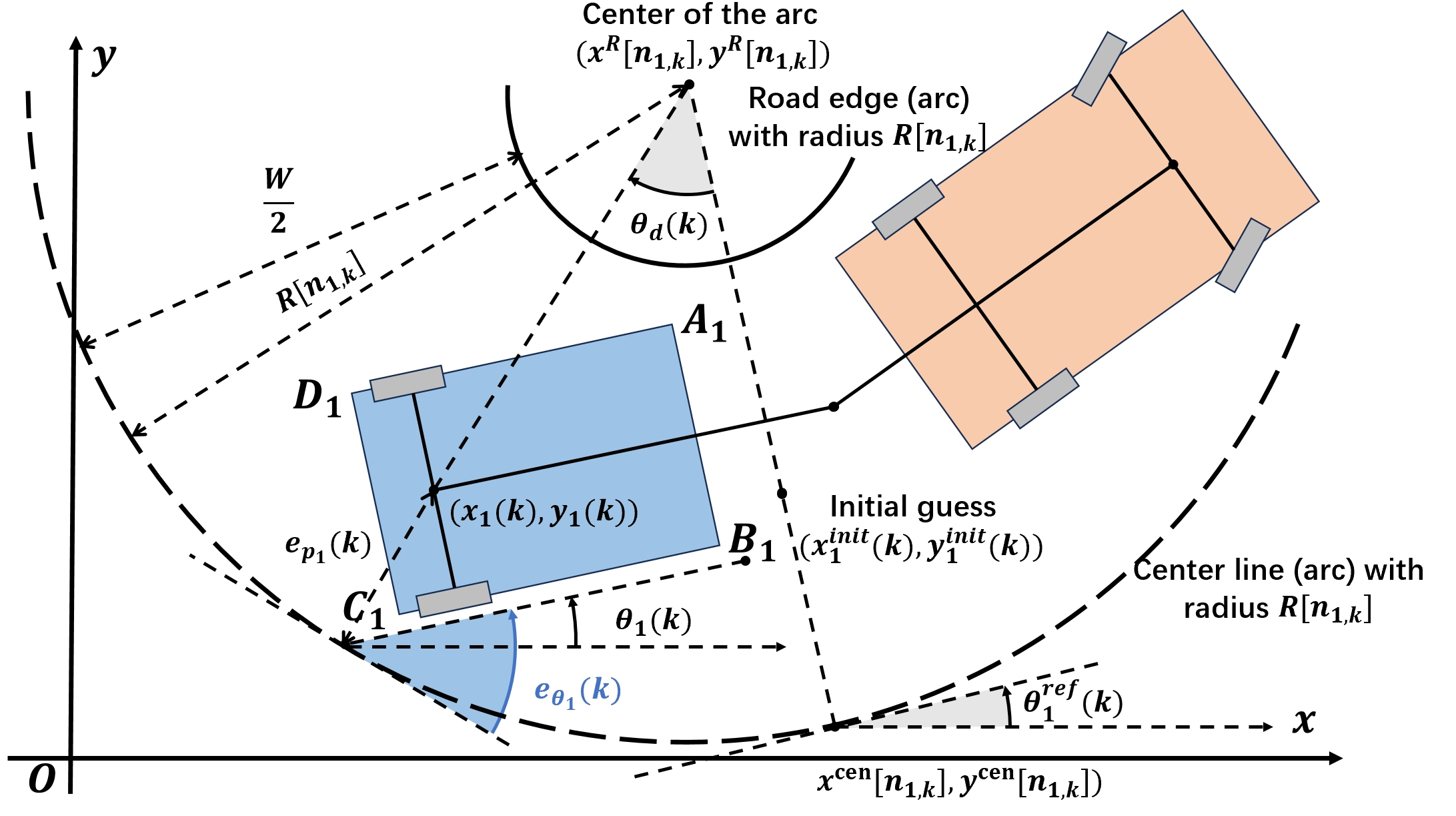}
\label{Center line is arc. Take the trailer as an example}}
\caption{The geometric formulation to characterize the lateral and orientation errors from center line. The reference point $(x_i^{\rm ref}, y_i^{\rm ref})$ is the projection of the initial guess point $(x_i^{\rm init}, y_i^{\rm init})$ on the center line. }
\label{Lateral and orientation error from center line}
\end{figure}
As shown in Fig. \ref{Center line is straight line. Take the tractor as an example}, when the projection is on a straight segment, the lateral and orientation errors shall take the form
\begin{equation}
\label{straight line}
\begin{aligned}
&e_{p_i}(k) = (y_i(k) - y^{\rm cen}[n_{i,k}])\cos\theta^{\rm cen}[n_{i,k}] \\ & \qquad \;\:\: - (x_i(k) - x^{\rm cen}[n_{i,k}])\sin\theta^{\rm cen}[n_{i,k}], \\
&e_{\theta_i}(k)  = \theta_i(k) - \theta^{\rm cen}[n_{i,k}], i, k \in \{i,k| R[n_{i,k}] = \infty\}, \\
\end{aligned}
\end{equation}
where $R[n_{i,k}]$ is the radius of the segment which the reference point $(x^{\rm cen}[n_{i,k}], y^{\rm cen}[n_{i,k}])$ belongs to, namely, $R[n_{i,k}] = \infty$ for straight line segment and $R[n_{i,k}] \neq \infty$ for arc segment.

On the other hand, when the projection is on an arc segment, as shown in Fig. \ref{Center line is arc. Take the trailer as an example}, the lateral and orientation errors are
\begin{equation}
\label{arc}
\begin{aligned}
& \!e_{p_i}(k) \!=\! \|x_i(k) \!-\! x^R[n_{i,k}], y_i(k) \!-\! y^R[n_{i,k}]\| \!-\! R[n_{i,k}],\\
& \!e_{\theta_i}(k) \!=\! \theta_i(k) \!-\! \theta^{\rm cen}\![n_{i,k}] \!+\! \theta_d(k),\! i,\! k \!\in\! \{i,\!k| R[n_{i,k}] \!\!\neq\!\! \infty\},\\
\end{aligned}
\end{equation}
where $(x[n_{i,k}], y[n_{i,k}])$ is the center point of the arc segment. $\theta_d$ is the angle difference between the variable and initial guess. The lateral and orientation errors in cost function \eqref{Cost Function} are analytically calculated through \eqref{straight line} and \eqref{arc}.

Next, the lateral and orientation errors shall compose the road edge constraints \eqref{eq: road edge constraints}. More specifically, on the one hand, as shown in Fig. \ref{Center line is straight line. Take the tractor as an example},  when the center line is a straight line, the distance of the vehicle's vertexes to the center line would be computed from $e_{p_i}(k)$ and $e_{\theta_i}(k)$. Therefore, the road edge constraints shall take the form:
\begin{equation}
\label{eq: straight line tractor-trailer feasible region constraints}
\begin{aligned}
&\pm e_{p_0}(k) + \frac{w_i}{2}\cos e_{\theta_0}(k) + l_{f_i}\sin e_{\theta_0}(k) \leq W/2,\\
&\pm e_{p_0}(k) + \frac{w_i}{2}\cos e_{\theta_0}(k) - l_{r_i}\sin e_{\theta_0}(k) \leq W/2,\\
&i, k \in \{i,k| R[n_{i,k}] = \infty\},\\
\end{aligned}
\end{equation}
where $W$ is the width of the road, $l_{f_i}/l_{r_i}$ is the front/rear overhang of the tractor/trailer. On the other hand, as shown in Fig. \ref{Center line is arc. Take the trailer as an example},  when the center line is an arc, the road edge constraints shall take the form
\begin{equation}
\label{eq: arc trailer-trailer vehicle feasible region constraints}
\begin{aligned}
& |\|x_{v_i}(k) \!-\! x^R[n_{i,k}], y_{v_i}(k) \!-\! y^R[n_{i,k}]\| \!-\! R[n_{i,k}]| \!\leq\! \frac{W}{2},\\
&v_i \in \{A_i, B_i, C_i, D_i\}, i, k \in \{i,k|R[n_{i,k}] \neq \infty\},\\
\end{aligned}
\end{equation}
where $(x_{v_i}(k), y_{v_i}(k)),  v_i \in \{A_i, B_i, C_i, D_i\}$ is the position of the tractor's/trailer's vertex at the $k$-th time-step.
In summary, \eqref{eq: straight line tractor-trailer feasible region constraints} and \eqref{eq: arc trailer-trailer vehicle feasible region constraints} give the explicit expression of the road edge constraints \eqref{eq: road edge constraints}.

\subsection{Generating a coarse trajectory as an initial guess}
\label{Coarse Trajectory Generation Section}
Next, we generate a coarse trajectory based on the center line to provide an intuitive initial guess for the subsequent optimization process.
The main idea of the generation process is dividing the center line into several ``wide'' and `narrow'' segments based on the collision risk. Then we generate a local path for each segment and connect them into a complete coarse path. Finally, we find a coarse trajectory from the coarse path.

More specifically, inspired by \cite{lian2023}, for each point $P^{\rm cen}[k]=(x^{\rm cen}[k], y^{\rm cen}[k], \theta^{\rm cen}[k])\in \mathcal{C}$, we regard the point to be ``wide" if there is no collision risk for any tractor/trailer pose configuration $\hat{P}_i:=(x_i,y_i,\theta_i), i=0,1$ that belongs to a neighbourhood around $P^{\rm cen}[k]$, and ``narrow" otherwise. 
Namely, given a center line point $P^{\rm cen}\in\mathcal{C}$ and obstacle set $\mbox{Obs}:=\{\mbox{Obs}_j\}_{j=1}^{N_{obs}}$, the collision risk of $P^{\rm cen}$ is computed by
\begin{equation}
\label{eq:check_collision_risk}
\begin{aligned}
    &\mbox{Risk}(P^{\rm cen})\!=\!\bigwedge_{\substack{\hat{P}_i\in\mathcal{U}(P^{\rm cen}), i=0,1, \\j=1,\ldots,N_{obs}}}\!
    \begin{aligned}
        &\mbox{OutPoly}(p,\mbox{Obs}_j), p \in \mbox{Veh}_i,\\
        &\mbox{OutPoly}(q,\mbox{Veh}_i), q \in \mbox{Obs}_j,\\
    \end{aligned}\\  
\end{aligned}
\nonumber
\end{equation}
where $\mbox{Risk}(P^{\rm cen})$ is denoted as the collision risk of the center line point $P^{\rm cen}$, $\mathcal{U}(P^{\rm cen})$ is the neighborhood around $P^{\rm cen}$, namely,
$\mathcal{U}(P^{\rm cen}) = \{\hat{P}_i: |x_i \!-\! x^{\rm cen}| \!\leq p_{\rm thre},|y_i \!-\! y^{\rm cen}| \leq p_{\rm thre},
|\theta_i \!-\! \theta^{\rm cen}| \leq \theta_{\rm thre}, \;i = 0,1\},$
where $p_{\rm thre}$ and $\theta_{\rm thre}$ are the position and orientation thresholds.
We then push the point into the ``wide"/``narrow" segment set $S_{PW}/S_{PN}$ according to its collision risk.  

\begin{figure}[!htbp]
\begin{center}
\subfigure[The process of finding local obstacles.]
{\includegraphics[width=0.45\hsize]{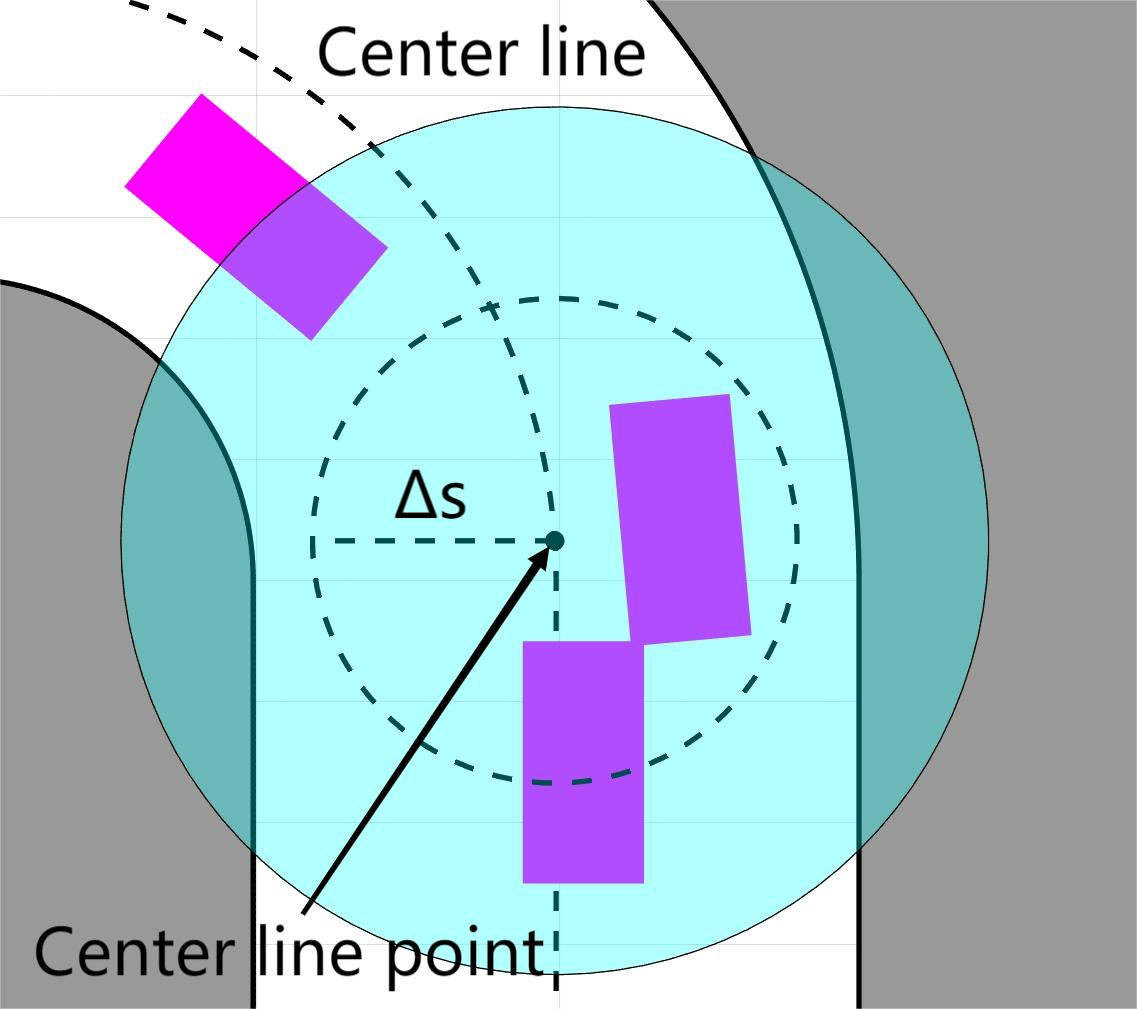}
\label{fig: The process of finding local obstacles}} 
\subfigure[The ``wide" and ``narrow" segments.]
{\includegraphics[width=0.45\hsize]{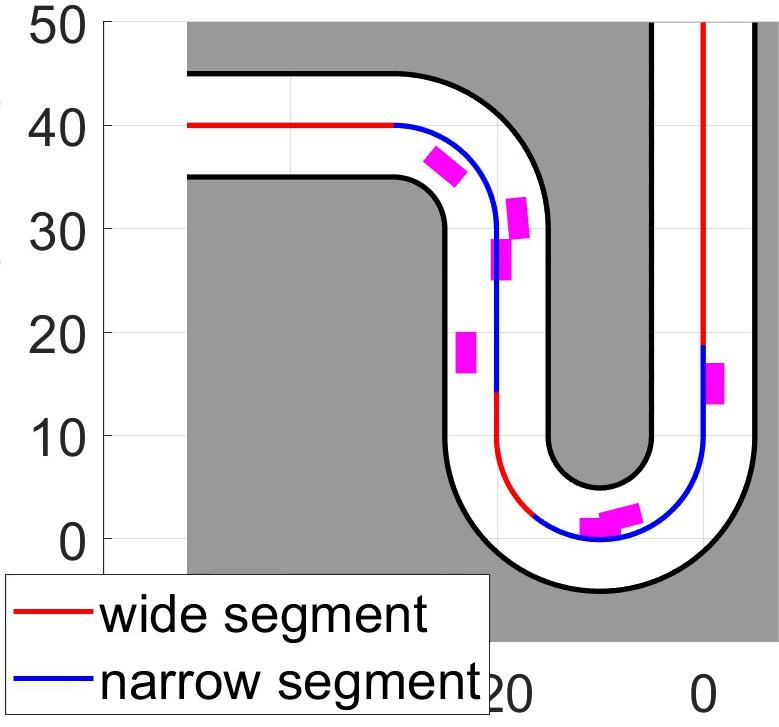}
\label{fig: The ``wide" and ``narrow" segments}} 
\caption{The process of generating coarse path.}
\label{fig: The process of generating coarse path.}
\end{center}
\end{figure}

On the other hand, as shown in Fig. \ref{fig: The process of finding local obstacles}, we find the local obstacles $\mathcal{O}[k,0]$ and $ \mathcal{O}[k,1]$ for the tractor and the trailer, respectively. More specifically, if ${\rm Obs}_j \in \mathcal{O}[k,i]$, it means that ${\rm Obs}_j$ has the collision risk when the tractor's/trailer's position is in the local area: $\|x_i(k)-x^{\rm cen}[n_{i,k}], y_i(k)-y^{\rm cen}[n_{i,k}]\| \leq \Delta s$. In the subsequent optimization process, when the tractor/trailer's position initial guess projection on the $k$-th center line point $P^{\rm cen}[k]$, we shall only consider the local obstacles that belong to $\mathcal{O}[k,0] / \mathcal{O}[k,1]$ by restricting the tractor's/trailer's position inside the local area. Consequently, the number of the collision-avoidance constraints would be significantly reduced.
 
By repeating such process, as shown in Fig. \ref{fig: The ``wide" and ``narrow" segments}, the center line set $\mathcal{C}$ is divided into ``wide'' $S_{SW}$ and ``narrow'' segment sets $S_{SN}$.  
Note that the $i$-th ``wide''/``narrow'' segment $\{(x[k_n], y[k_n], \theta[k_n])\}_{n=0}^{N_{s_i}}$ is a subset of $\mathcal{C}$ with $N_{s_i}$ elements. 
We choose the center line segment to be the local coarse path of a road segment if the segment is ``wide"  
\begin{align*}
    {\rm LocPath}[i]\!=\!\{x[k_n], y[k_n], x[k_{n-N_{\delta}}], y[k_{n-N_{\delta}}]\}_{n = N_{\delta}}^{N_{s_i}},
\end{align*}
where $(x[k_n], y[k_n])$ and $(x[k_{n-N_{\delta}}], y[k_{n-N_{\delta}}])$ are the path points of the tractor and the trailer, respectively; $N_{\delta} := \lfloor (l_H + l_1)/{\delta s} \rfloor$ is the number of the offset points between the tractor and the trailer. When the road segment is ``narrow", we use A* algorithm to plan the local coarse paths for the tractor and the trailer, respectively. Finally, we connect all the local coarse paths sequentially into a global path $\hat{\mathcal{P}}:=\cup_{i}\mbox{LocalPath}[i]$. 

Next, we find an initial guess trajectory from the resampled coarse path $\hat{\mathcal{P}}$. To this end, we first spatially down-sample the global coarse path $\hat{\mathcal{P}}$ to a path $\mathcal{P} := \{x_{0}^{\rm init}[k],$$ y_{0}^{\rm init}[k],x_{1}^{\rm init}[k], y_{1}^{\rm init}[k]\}_{k=0}^{N_{\rm curr}}$ with $N_{\rm curr}$ points. 
Note that the resampled coarse path $\mathcal{P}$ only contains the positions of the tractor and the trailer, it is insufficient to serve as the initial guess. Hence we generate $\{\theta_{0}^{\rm init}(k), $ $\theta_{1}^{\rm init}(k), \varphi^{\rm init}(k),v^{\rm init}(k)\}_{k=0}^{N_{\rm curr}}$ in a similar way to \cite{lian2023}. 
\begin{algorithm}[!htb]
  \label{Coarse Trajectory Generation}
  \caption{Coarse Trajectory Generation}
  \SetAlgoLined
  \KwIn{Obstacles Obs, and Center Line Points Set $\mathcal{C}$;}
  \KwOut{Coarse trajectory $\mathcal{T}^{\rm init}$, and Local Obstacle Set $\mathcal{O}$;}
  Wide Segment Set $S_{SW} \leftarrow \emptyset $;  Wide Point Set $S_{PW}$ $ \leftarrow \emptyset$;  Narrow Point Set $S_{PN} \leftarrow \emptyset$; Narrow Segment Set $S_{SN} \leftarrow \emptyset$;  
  Local Obstacle Set $\mathcal{O} \leftarrow \emptyset$\;
  \For{\rm {each point} $P^{\rm cen}[k] \in \mathcal{C}$}
  {      
  $\mathcal{O}[k, 0], \mathcal{O}[k, 1] \leftarrow$ FindLocObstacle$(P^{\rm cen}[k], {\rm Obs})$\;
  \uIf {{\rm Risk}$(P^{\rm cen}[k]) = 0$}
      {push $P^{\rm cen}[k]$ into $S_{PW}$\;
      \If{$S_{PN} \neq \emptyset$} 
      {push $S_{PN}$ into $S_{SN}$; $S_{PN} \leftarrow \emptyset$\;} 
      }
  \Else
      {push $P^{\rm cen}[k]$ into $S_{PN}$\;
      \If{$S_{PW} \neq \emptyset$} 
      {push $S_{PW}$ into $S_{SW}$; $S_{PW} \leftarrow \emptyset$\;}
      }
  }
${\rm LocPath} \leftarrow $PlanLocalPaths$(S_{SW},S_{SN})$\;
$\hat{\mathcal{P}} \leftarrow $ConnectLocalPaths$({\rm LocPath})$\;
$\mathcal{P}\leftarrow$ ResamplePath$(\hat{\mathcal{P}}, N_{\rm curr})$\;
$\mathcal{T}^{\rm init}$ $\leftarrow $ ConvertToTraj$(\mathcal{P})$\;
\Return{$\mathcal{T}^{\rm init}, \mathcal{O}$}
\end{algorithm}

The pseudo code of the coarse trajectory generation algorithm is summarized in Alg. \ref{Coarse Trajectory Generation}. 
Therein, the function FindLocObstacle$(P^{\rm cen}[k], {\rm Obs})$ finds the local obstacles $\mbox{LocObs}[k]$;
PlanLocalPaths$(S_{SW},S_{SN})$ plans a local coarse path for each segment; ConnectLocalPaths$({\rm LocPath})$ connects the local path sequentially; ResamplePath$(\hat{\mathcal{P}},N_{\rm curr})$ spatially down-sample the global coarse path $\hat{\mathcal{P}}$ to $N_{\rm curr}$ points; and ConvertToTraj$(\mathcal{P})$ get the coarse trajectory $\mathcal{T}^{\rm init}$ from $\mathcal{P}$.

\subsection{The Progressively Increasing Sampling Number Optimization (PISNO) framework}

In this section, we propose the PISNO framework to solve the trajectory planning problem efficiently using the coarse trajectory obtained from \ref{Coarse Trajectory Generation Section}.
The PISNO framework comprises two steps. In step 1, we find a nearly feasible trajectory with a small sampling number. In step 2, we sequentially solve a series of simplified OCP with a warm-starting technique that progressively increases the sampling number up to the threshold sampling number $N_{\rm thre}$. 

\subsubsection{Finding a nearly feasible trajectory}
The objective of this step is to plan a nearly feasible trajectory from the initial guess $\mathcal{T}^{\rm init}$.
Inspired by \cite{li2021tits}, we first soften the nonlinear kinematic constraints \eqref{kinematic model of tractor-trailer} and \eqref{trailer position} by adding the violation of the kinematic constraints into the cost function. In addition, we replace the large-scaled triangle-area-based constraints \eqref{Collision-Avoidance Constraints} and road edge constraints \eqref{eq: straight line tractor-trailer feasible region constraints}, \eqref{eq: arc trailer-trailer vehicle feasible region constraints} with the linear ``within-corridor constraints" introduced in \cite{li2021tits}.
Consequently, the OCP shall take the form
\begin{equation}
\label{warm-start OCP Formulation}
\tag{OCP$_{\rm{warm}}$}
\begin{aligned}
\min_{\substack{\bm{x},\bm{u}}}\;\;\;\; & \mbox{Cost function } \eqref{Cost Function} + \omega_{\rm penalty}\cdot J_{\rm penalty}, \\
\mbox{s.t.\;\;\;\;} & \rm Kinematic \ constraints \ \eqref{state and control bounds},\\
& \rm Boundary \ constraint \  \eqref{Boundary Constraint}, \\
& \rm Within\mbox{-}corridor \ constraints ,\\
\end{aligned}
\end{equation}
where $J_{\rm penalty}$ is the penalty for the violation of kinematic and $\omega_{\rm penalty}$ is the weight of the penalty. More specifically, ``within-corridor constraints" find the collision-free convex polygons around the initial guess. Here we use the same formulation of $J_{\rm penalty}$ and ``within-corridor constraints" as \cite{li2021}. The detailed expressions are omitted for brevity.

\eqref{warm-start OCP Formulation} is solved efficiently due to the fact that: (i) all of the constraints are linear and the only nonlinearity lies in the objective function; (ii) the sampling number $N_{\rm curr}$ is relatively small.
Nevertheless, a too big penalty weight $\omega_{\rm penalty}$ jeopardizes the goal of minimizing the original cost function \eqref{Cost Function}.
and the solution is far from kinematically feasible if $\omega_{\rm penalty}$ is not large enough. 
Therefore, kinematic constraints \eqref{kinematic model of tractor-trailer} and \eqref{trailer position} should be resumed in the subsequent optimization process.
In addition, the ``within-corridor constraints" discard some feasible regions \cite{li2021}. Thus the numerical optimization solver will return a sub-optimal solution, or even fail to find a feasible solution if we resume the kinematic constraints. 
We hence resume the triangle-area-based collision-avoidance criterion \cite{li2015} later to retain more collision-free space.

\subsubsection{Progressively increasing the sampling number}
Step 2 aims to find an optimal trajectory using the nearly feasible trajectory obtained from step 1. Instead of directly solving \eqref{Overall Formulation} with the threshold sampling number $N_{\rm thre}$, we iteratively solve a sequence of intermediate OCPs with the sampling number gradually increases to $N_{\rm thre}$. In each iteration, the sampling number is increased by a multiplier $\alpha$. Then, the trajectory is resampled temporally by imposing a linear interpolation for the trajectory. The resampled trajectory with increased sampling number would serve as the initial guess for the subsequent intermediate OCP. 
Then, the subsequent intermediate OCP is solved and the iteration starts over again if $N_{\rm curr} < N_{\rm thre}$.

On the other hand, we redesign the collision-avoidance constraints based on the type of the reference points (``wide'' or ``narrow'') to reduce the number of collision-avoidance constraints.
More precisely, given the initial guess $\{P_i^{\rm init}(k) := (x_{i}^{\rm init}(k),y_{i}^{\rm init}(k), $ $ \theta_{i}^{\rm init}(k)), i \!\in\! \{0,1\}\}_{k=0}^{N_{\rm curr}}$, we compute the projections of the tractor and the trailer 
$ \{P^{\rm cen}[n_{i,k}] = (x^{\rm cen}[n_{i,k}],$ $ y^{\rm cen}[n_{i,k}], \theta^{\rm cen}[n_{i,k}]), i \in \{0,1\} \}_{k=0}^{N_{\rm curr}}=:\mathcal{R}$.
If the type of $P^{\rm cen}[n_{i,k}]$ is ``wide'' and the initial guess point $P_i^{\rm init}(k) \in \mathcal{U}(P^{\rm cen}[n_{i,k}])$, we guarantee the safety by keeping the tractor/trailer's pose  $P_i(k) \in \mathcal{U}(P^{\rm cen}[n_{i,k}])$. Thus, the constraints in this case take the form
\begin{equation}
\label{wide segment collision-avoidance}
\begin{aligned}
&P_i(k)\in\mathcal{U}(P^{\rm cen}[n_{i,k}]), \\
&i,\!k \!\in\! \{i,\!k|P^{\rm cen}[n_{i,k}] \!\in\! S_{SW} \!\wedge\! P_i^{\rm init}\!(k) \!\in\! \mathcal{U}(P^{\rm cen}[n_{i,k}])\}.\\
\end{aligned}
\end{equation}
where $P_i(k) := (x_i(k), y_i(k), \theta_i(k))$ is the pose of the tractor/trailer. 
Otherwise, recall that we have characterized the local obstacle set $\mathcal{O}$ in Section \ref{Coarse Trajectory Generation Section}, we only need to consider the collision-avoidance with the local obstacles that belong to $\mathcal{O}[n_{i,k}, i]$. Namely,
\begin{equation}
\label{narrow segment collision-avoidance}
\begin{aligned}
    &\|x_i(k)-x^{\rm cen}[n_{i,k}], y_i(k)-y^{\rm cen}[n_{i,k}]\| \leq \Delta s,\\
    &{\rm OutPoly}(p, \mbox{Obs}_j), p \in \mbox{Veh}_i(k), \\
    &{\rm OutPoly}(q, \mbox{Veh}_i(k)), q \in \mbox{Obs}_j, j \in \mathcal{O}[n_{i,k}, i],\\
    &i,\!k \!\in\! \{i,\!k|P^{\rm cen}[n_{i,k}] \!\in\! S_{SN} \!\vee\! P_i^{\rm init}\!(k) \!\notin\! \mathcal{U}(P^{\rm cen}[n_{i,k}])\}.\\
\end{aligned}
\end{equation}
where $\Delta{s}$ is the radius of the local area. Thus, the intermediate simplified OCP with sampling number $N_{\rm curr}$ takes the form.
\begin{equation}
    \label{simplified OCP}
    \tag{OCP$_{\rm simplified}$}
    \begin{aligned}
    \min_{\substack{\bm{x},\bm{u}}}\;\;\;\; &\mbox{Cost function } \eqref{Cost Function}, \\
    \mbox{s.t.\;\;\;\;\;} & \rm Kinematic \ constraints \ \eqref{kinematic model of tractor-trailer}, \eqref{trailer position}, \eqref{state and control bounds}, \\
    &\rm Boundary \ constraints \  \eqref{Boundary Constraint}, \\
    &\rm Size\mbox{-}reduced \ collision\mbox{-}avoidance \\
    & \rm constraints \ \eqref{wide segment collision-avoidance}, \eqref{narrow segment collision-avoidance},\\
    &\rm Road \ edge \ constraints \ \eqref{eq: straight line tractor-trailer feasible region constraints}, \eqref{eq: arc trailer-trailer vehicle feasible region constraints}.
    \end{aligned}
\end{equation}
Note that, the number of the constraints in \eqref{wide segment collision-avoidance} and \eqref{narrow segment collision-avoidance} are much less than that in \eqref{Collision-Avoidance Constraints} because the number of \eqref{wide segment collision-avoidance} is not related to the number of the obstacles $N_{\rm obs}$ and the number of \eqref{narrow segment collision-avoidance} is only related to number of the local obstacles.
Moreover, \eqref{simplified OCP} could be solved further efficiently due to the extra facts that: (i) the sampling number $N_{\rm curr} < N_{\rm thre}$ except for the final iteration; (ii)
the solution of \eqref{simplified OCP} in the previous iteration warm-starts the next iteration.

\begin{figure}[!htbp]
\centerline{\includegraphics[width=1\linewidth]{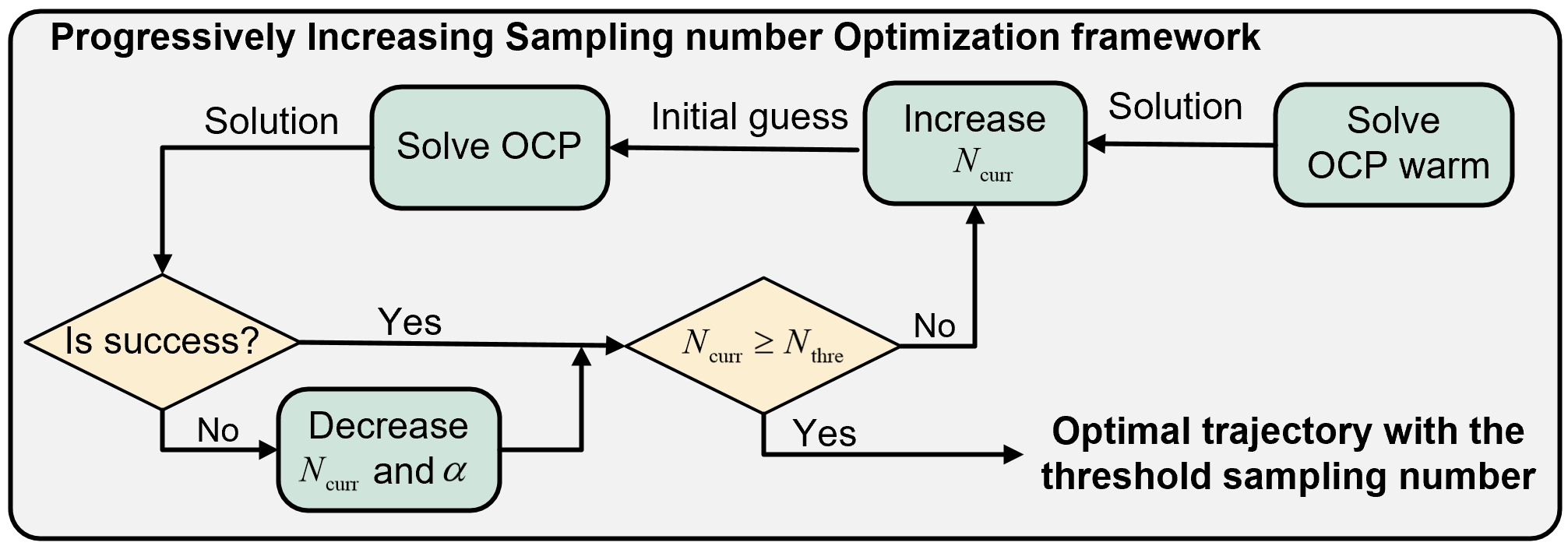}}
\caption{The flow diagram of the proposed PISNO framework.}
\label{fig: diagram}
\end{figure}

To give an overview of the proposed PISNO framework, we illustrate the procedure with a flow diagram in Fig. \ref{fig: diagram}, and the complete pseudo-code is summarized in Algorithm \ref{Alg: Progressively Increasing Sampling Number Optimization (PISNO) Method}.
\begin{algorithm}
  \label{Alg: Progressively Increasing Sampling Number Optimization (PISNO) Method}
  \caption{PISNO Framework}
  \SetAlgoLined
  \KwIn{Local Obstacles Set $\mathcal{O}$, Center Line Points Set $\mathcal{C}$, and Coarse trajectory $\mathcal{T}^{\rm init}$;}
  \KwOut{Optimized trajectory $\mathcal{T}$;}
  \textbf{Step 1: Finding a nearly feasible trajectory} \\
  $\mathcal{R} \leftarrow$ FindRefPoints$(\mathcal{T}^{\rm init}, \mathcal{C})$\;
  $\rm OCP_{warm} \leftarrow $ 
  FormulateWarmOCP$({\rm \mathcal{T}^{\rm init}}, \mathcal{R})$\;
  $\mathcal{T} \leftarrow$ SolveOCP($\rm OCP_{warm}$)\;
  \textbf{Step 2: Progressively increasing sampling number} \\
  \While{$N_{\rm curr} < N_{\rm thre}$}
  {$N_{\rm curr} \leftarrow \min(N_{\rm thre}, \alpha \cdot N_{\rm curr})$\; 
  $\mathcal{T} \leftarrow$ ResampleTraj$(\mathcal{T}, N_{\rm curr})$\;
  $\mathcal{R} \leftarrow$ FindRefPoints$(\mathcal{T}, \mathcal{C})$ \;
  ${\rm OCP_{simplfied}} \leftarrow $ FormulateOCP$(\mathcal{T}, \mathcal{R}, \mathcal{O})$\;
  \uTry{$\mathcal{T} \leftarrow$ 
  SolveOCP(${\rm OCP_{simplfied}}$)\;}
  \uCatch{
    $N_{\rm curr} \leftarrow N_{\rm curr} / \alpha $\;
    $\alpha \leftarrow \alpha - \gamma$\;
    \If{$\alpha \leq 1$}
    {\Return $\emptyset$\;}
  }
  }
  \Return $\mathcal{T}$\;
\end{algorithm}
Therein, the function FindRefPoints$(\cdot,\mathcal{C})$ finds the projection points set $\mathcal{R}$; ResampleTraj$(\mathcal{T}, N_{\rm curr})$ resamples the trajectory temporally into $N_{\rm curr}$ points; FormulateWarmOCP$(\mathcal{T}^{\rm init}, \mathcal{R})$ formulates \eqref{warm-start OCP Formulation} and FormulateOCP$(\mathcal{T}, \mathcal{R}, \mathcal{O})$ formulates \eqref{simplified OCP} with the current sampling number $N_{\rm curr}$; SolveOCP($\cdot$) solves \eqref{warm-start OCP Formulation}/\eqref{simplified OCP}.
\begin{remark}
\label{remark idea}
The key idea of the proposed PISNO framework in Algorithm 2 is that we use the optimal trajectory with a
small sampling number as the initial guess to warm-start the OCP with a large sampling number. As the the initial guess is the near-optimal solution of the OCP with a large sampling number, the numerical optimization process will be significantly accelerated. 
\end{remark}
\begin{remark}
\label{remark limitation}
If the algorithm fails in solving the intermediate OCPs, the possible reasons include: (i) the solution of the last iteration is far from being feasible for the OCP, and (ii) the scenario is too narrow to exist a feasible solution. For the first failure reason, we would try to solve the intermediate OCPs with a milder sampling number increasing process by reducing the multiplier $\alpha$.
\end{remark}

\section{Simulation and experiment results}
\subsection{Simulation results}
We implement our algorithm in Matlab 2022b and execute it on an i5-12500H CPU with 16GB RAM. Regarding the solver in function SolveOCP($\cdot$), we choose the interior-point solver IPOPT \cite{IPOPT} and set the linear solver as ``MA27" \cite{MA27} in AMPL \cite{AMPL}. 
Parametric settings are listed in Table \ref{Parametric settings}.
\begin{table}[!htbp]
\caption{Parametric settings in simulation}
\begin{center}
\begin{tabular}{cccc}
\toprule
Parameter & Settings & Parameter & Settings\\
\midrule
$T$ & 20s & $l_0$ & $2m$ \\
$l_H$ & $1m$ & $l_1$ & $4m$ \\
$w_0$ & $2m$ & $w_1$ & $2m$ \\
$l_{f_0}$ & $3m$ & $l_{r_0}$ & $1m$ \\
$l_{f_1}$ & $3m$ & $l_{r_1}$ & $1m$ \\
$v_{max}$ & $5m/s$ & $\varphi_{max}$& $0.7rad$ \\
$a_{max}$ & $5m/s^2$ & $\omega_{max}$ & $1rad/s$\\
$\delta\theta_{max}$ & $1 rad$ & $\delta s$ & $0.5m$\\
$\omega_g$ & $1$ & $\omega_{e}$ & $1$\\
$\omega_c$ & $10$ & $\omega_{\theta}$ & $2$\\
$p_{\rm thre}$ & $1m$ & $\theta_{\rm thre}$ & $0.2rad$\\
$W$ & $10m$ & $\Delta s$ & $4m$\\
$\alpha$ & $2$ & $\gamma$ & $0.5$\\
$N_{\rm curr}$ & $25$ & $N_{\rm thre}$ & $200$\\
\bottomrule
\end{tabular}
\label{Parametric settings}
\end{center}
\end{table}

To evaluate our trajectory planning algorithm, we simulate a challenging driving environment that includes a right turn and a U-turn, the static vehicles are set as obstacles around the center line. We compare our method with other sampling-and-search-based and optimization-based methods. In particular, extended hybrid A* (EHA) algorithm \cite{li2019} is used as a representative for the sampling-and-search-based methods. 
To make the comparison fair, we enhance EHA based on the on-road structure, similar to \cite{li2021}.
Regarding the comparison with the optimization-based planners, we choose Progressively Constrained Optimal Control (PCOC) method \cite{li2019}, Lightweight Iterative Optimization Strategy (LIOS) + Trust-Region-based Maneuver Optimization
(TRMO) method \cite{li2021}, Safety Travelling Constraints (STC) method \cite{cen2021} and the Adaptive Gradient-Assisted Particle
Swarm Optimization (AGAPSO) method \cite{chai2022multiphase} as the benchmarks. In particular, STC uses ``within-corridor constraints" to replace the collision-avoidance constraints in \eqref{Overall Formulation}. This would shrink the collision-free space and reduce the success rate of trajectory planning.
Therefore, to make a fair comparison, we enhance STC by adding a LIOS module. 

\begin{figure}[!htbp]
\centering
\subfigure[$N_{\rm curr}$ = 25]
{\includegraphics[width=0.48\hsize]{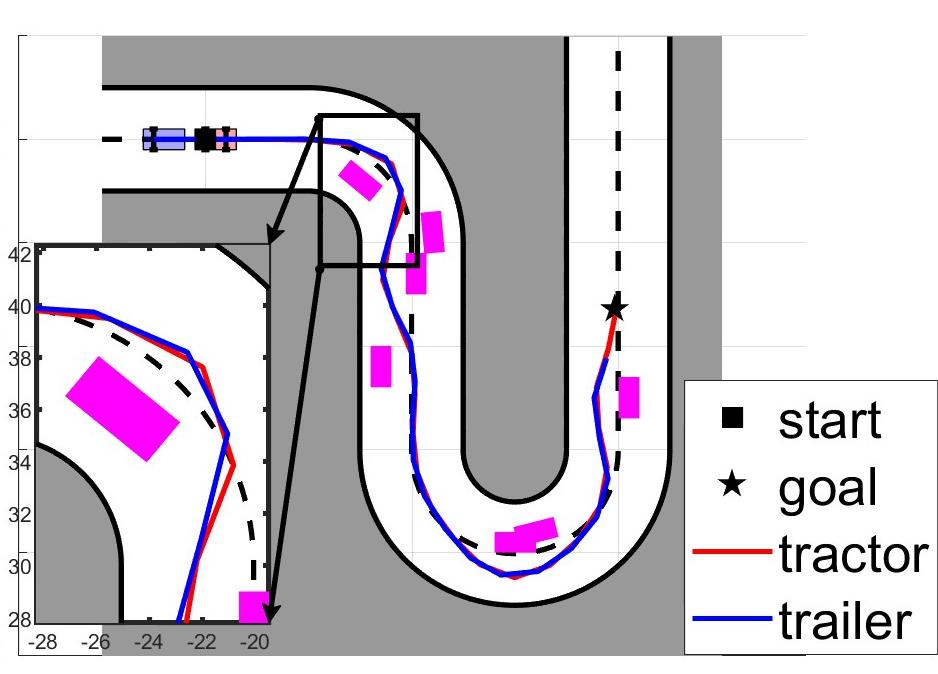}
\label{NC=25}}
\subfigure[$N_{\rm curr}$ = 50]
{\includegraphics[width=0.48\hsize]{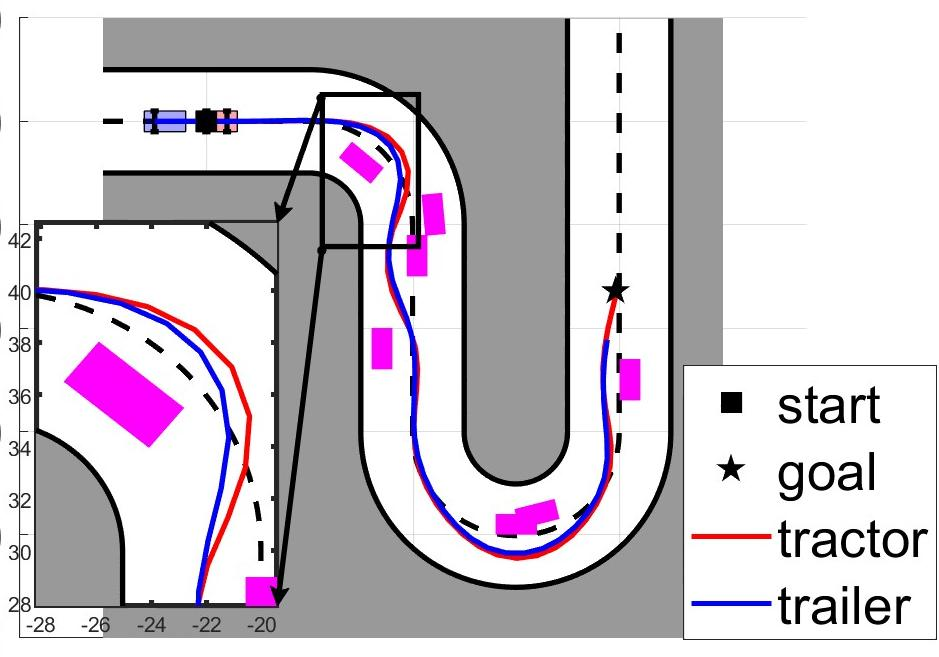}
\label{NC=50}}
\subfigure[$N_{\rm curr}$ = 100]
{\includegraphics[width=0.48\hsize]{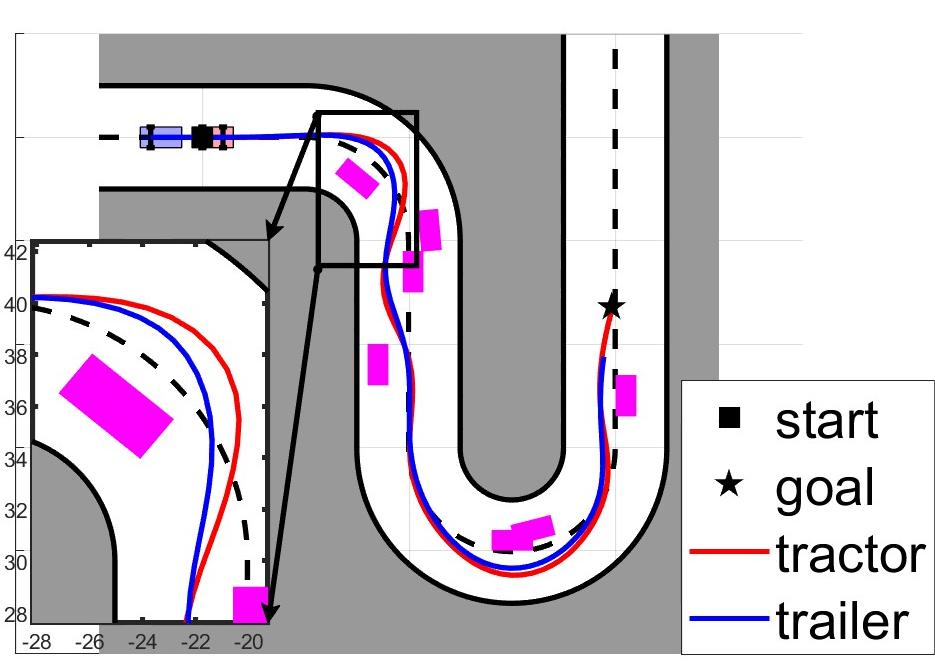}
\label{NC=100}}
\subfigure[$N_{\rm curr}$ = 200]
{\includegraphics[width=0.48\hsize]{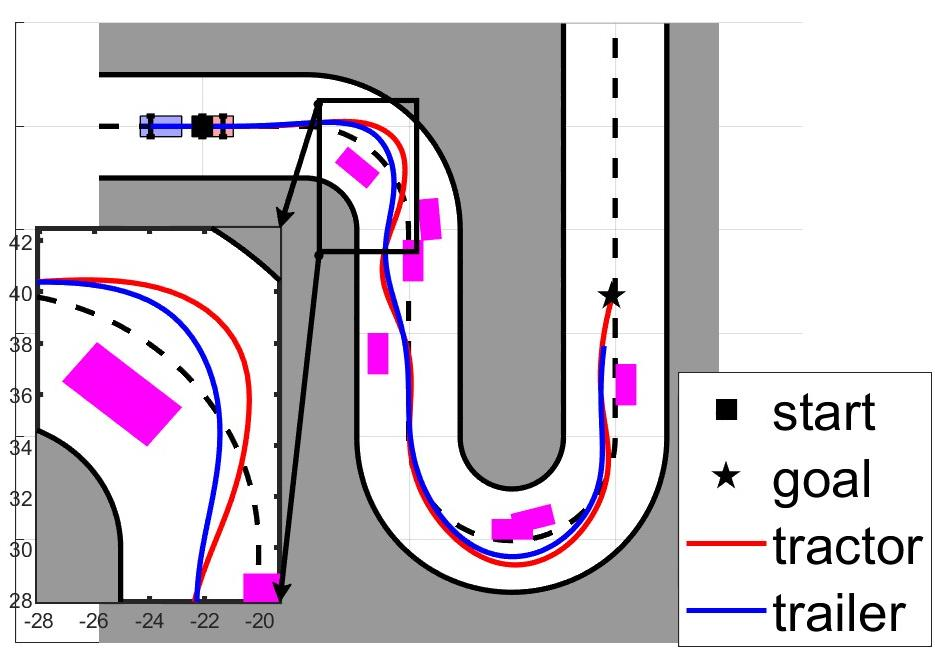}
\label{NC=200}}
\caption{The optimization process starts with a sampling number $N_{\rm curr} = 25$, and then refines the trajectory by progressively increasing the sampling number until $N_{\rm curr} = 200$ in (c)-(f).}
\label{``wide'' and ``narrow'' segments, coarse trajectory and intermediate solutions}
\end{figure}

We first evaluate the performance of our method with the other methods in the scenario with seven manually placed static obstacles to represent a congested road. 
The process of our planner is illustrated in Fig. \ref{``wide'' and ``narrow'' segments, coarse trajectory and intermediate solutions}. 
In Fig. \ref{NC=25}-Fig. \ref{NC=200}, as the sampling number $N_{\rm curr}$ increases, the planned trajectory gets smoother. 
\begin{figure}[!htbp]
\begin{center}
\subfigure[The profiles of velocity and steering angle.]
{\includegraphics[width=0.48\hsize]{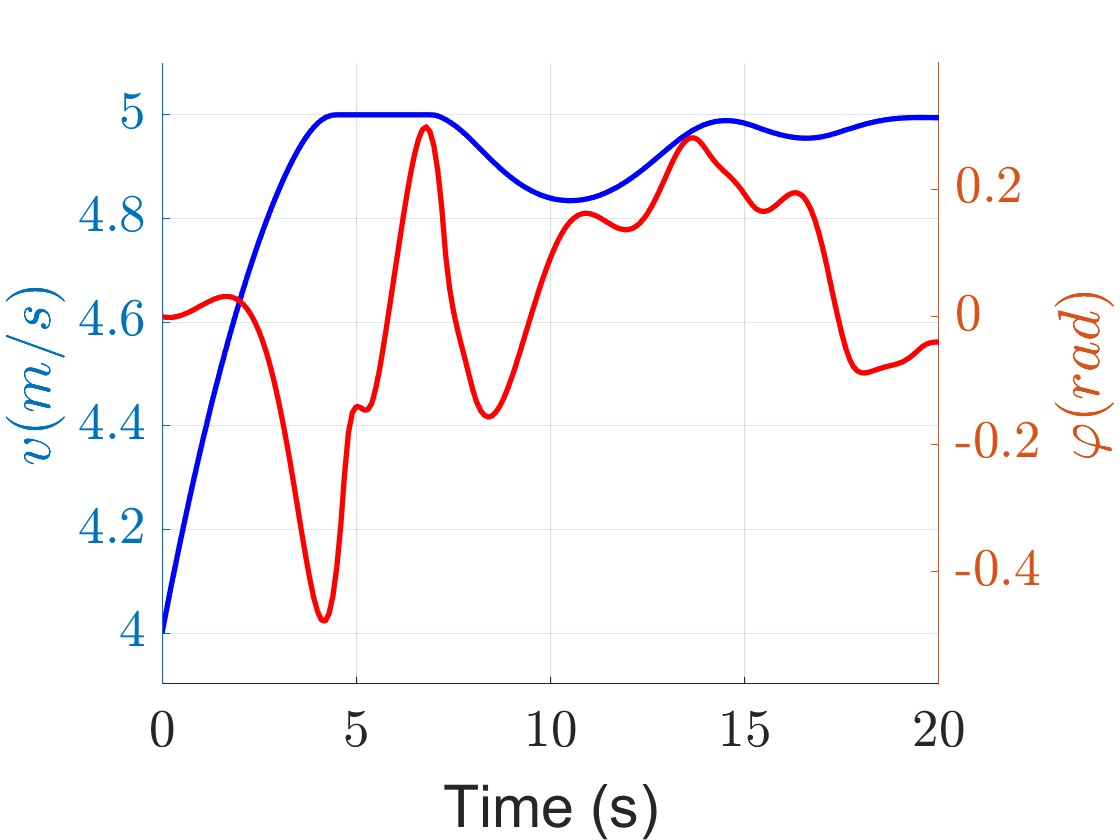}
\label{fig: The profiles of velocity and steering angle.}} 
\subfigure[The profiles of acceleration and steering angular velocity.]
{\includegraphics[width=0.48\hsize]{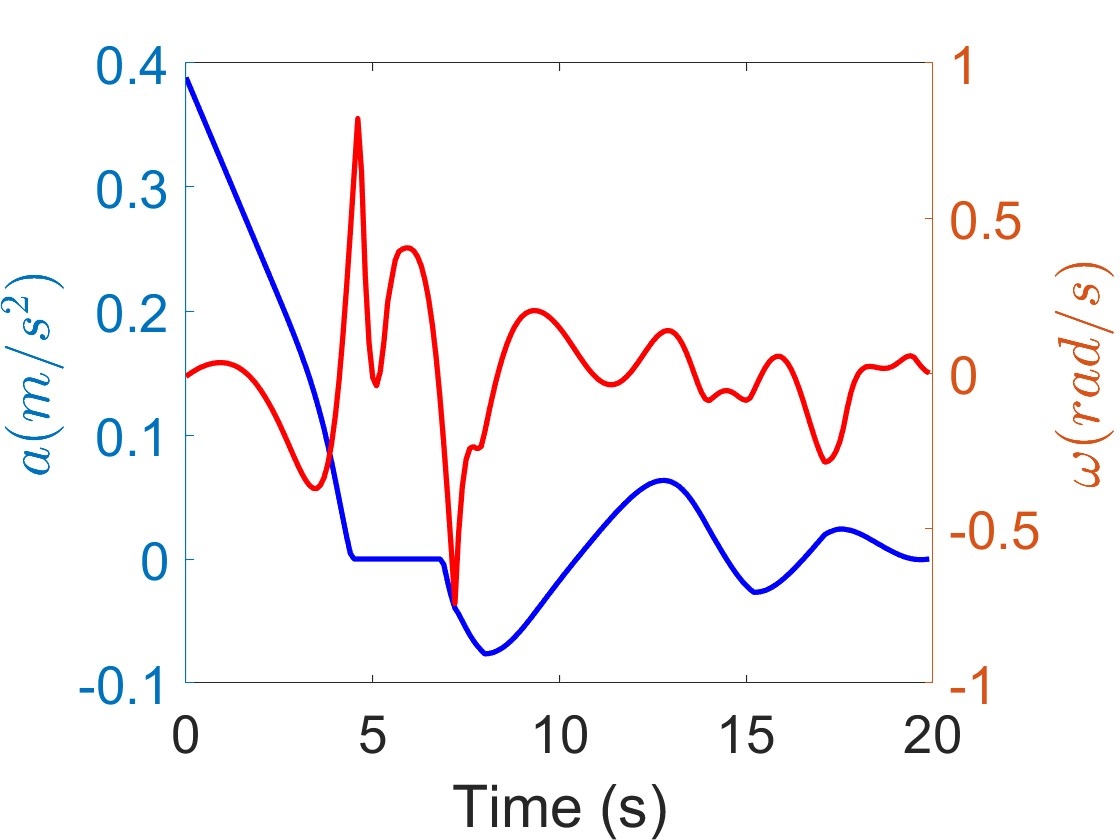}
\label{fig: The profiles of acceleration and steering angular velocity.}} 
\caption{The profiles of the vehicle state.}
\label{fig: The profiles of the vehicle state.}
\end{center}
\end{figure}
 
The velocity, steering angle, acceleration, and steering angular velocity profiles of the optimized trajectory are depicted in Fig. \ref{fig: The profiles of the vehicle state.}. The profiles highlight the smoothness of the movement, reflecting the trajectory's continuity and the gentle variation in controls.

\begin{table}[!htbp]
\caption{Comparison of different methods in a manually defined case}
\label{Comparison of different methods in a manually defined case}
\rowcolors{2}{gray!30}{white}
\begin{center}
\begin{tabular}{l|l|l}
\rowcolor{blue!10}
\textbf{Methods} & \textbf{Solving time(s)} & \textbf{Cost}\\
\hline
PISNO (Ours) & \textbf{2.58} & 68.71 \\
AGAPSO\cite{chai2022multiphase} & 11.90 & 67.87\\
EHA\cite{li2019} & 3.62 & \textbackslash \\
PCOC\cite{li2019} & 54.82 & \textbf{67.75} \\
LIOS + TRMO\cite{li2021} & 5.46 & 73.67 \\
STC\cite{cen2021} & 6.66 & 94.65 \\
\end{tabular}
\end{center}
\end{table}
We now compare the solving time of our method with the benchmarks. The results are illustrated in Table \ref{Comparison of different methods in a manually defined case}. 
It turns out that EHA, PCOC and STC are more time-consuming than our method. 
In particular, EHA becomes less efficient in narrow environment like Fig. \ref{``wide'' and ``narrow'' segments, coarse trajectory and intermediate solutions}.
And PCOC and STC are more time consuming since they are both based on EHA. 
Moreover, our method has the shortest solving time, this is mainly due to the fact that PISNO solves the intermediate OCP with a smaller sampling number except for the final iteration.
The proposed warm-start technique in PISNO further accelerates the solving process.

We further compare the optimal value of \eqref{Cost Function} of our method with those of the benchmarks. We do not compare the cost of EHA as it aims to find a feasible trajectory rather than an optimal one. STC results in the highest cost among the optimization-based methods because the ``within-corridor constraints'' are more strict than \eqref{Collision-Avoidance Constraints}. The cost of PISNO is slightly higher than that of PCOC. This is because the step of reducing the collision-avoidance constraints is equivalent to limiting the tractor-trailer's pose to a local feasible area. This potentially leads to a constrained feasible set.  

Furthermore, we conduct tests in 200 randomized scenarios to evaluate the robustness of the compared methods. In the first 100 random scenarios, to see the algorithms' performance under normal traffic conditions, we randomly place 6 to 9 obstacles in the simulation environment. In the second 100 random scenarios, we randomly place 12 to 15 obstacles in the simulated environment to check the algorithm's performance under dense traffic conditions. When generating the test cases, we remove the scenarios which do not allow for a feasible A* path. The results are shown in Table \ref{Comparison of different methods in random normal environments} and \ref{Comparison of different methods in random dense environments}.
\begin{table}[!htbp]
\caption{Comparison of different methods in random normal environments}
\label{Comparison of different methods in random normal environments}
\rowcolors{2}{white}{gray!30}
\begin{center}
\begin{tabular}{l|l|l|l|l}
\hline
\rowcolor{blue!10}
\textbf{Methods} & \textbf{Success}  & \textbf{Mean CPU}& \textbf{Max CPU} & \textbf{Mean}\\
\rowcolor{blue!10}
& \textbf{rate(\%)} & \textbf{time(s)} & \textbf{time(s)} & \textbf{Cost}\\
\hline
PISNO (Ours) & \textbf{100} & \textbf{2.25} & \textbf{2.79} & \textbf{37.65} \\
AGAPSO\cite{chai2022multiphase} & 97 & 13.69 & 71.93 & 41.06 \\
EHA\cite{li2019} & \textbf{100} & 3.59 & 10.65 & \textbackslash\\
PCOC\cite{li2019} & 97 & 33.74 & 175.91 & 45.76\\
LIOS + TRMO\cite{li2021} & \textbf{100} & 5.76  & 9.10 & 37.73\\
STC\cite{cen2021} & 82 & 8.24 & 19.03 & 56.31\\
\hline
\end{tabular}
\end{center}
\end{table}

\begin{table}[!htbp]
\caption{Comparison of different methods in random dense environments}
\label{Comparison of different methods in random dense environments}
\rowcolors{2}{white}{gray!30}
\begin{center}
\begin{tabular}{l|l|l|l|l}
\rowcolor{blue!10}
\hline
\textbf{Methods} & \textbf{Success}  & \textbf{Mean CPU}& \textbf{Max CPU} & \textbf{Mean}\\
\rowcolor{blue!10}
& \textbf{rate(\%)} & \textbf{time(s)} & \textbf{time(s)} & \textbf{Cost}\\
\hline
PISNO (Ours) & \textbf{99} & \textbf{3.24} & \textbf{6.30} & \textbf{45.84} \\
AGAPSO\cite{chai2022multiphase} & 90 & 33.94 & 112.25 & 118.71 \\
EHA\cite{li2019} & 97 & 7.93 & 54.46 & \textbackslash\\
PCOC\cite{li2019} & 95 & 114.45 & 359.48 & 121.45\\
LIOS + TRMO\cite{li2021} & 98 & 7.06  & 24.92 & 46.59\\
STC\cite{cen2021} & 67 & 10.35 & 54.45 & 62.18\\
\hline
\end{tabular}
\end{center}
\end{table}
As illustrated in Table \ref{Comparison of different methods in random normal environments}, PISNO, EHA and LIOS + TRMO methods all manage to plan trajectories in all 100 random scenarios under normal traffic conditions. STC achieves lower success rate and higher cost than PISNO and LIOS + TRMO methods due to the usage of the ``within-corridor constraints". Moreover, compared with all of the optimization-based methods, PISNO stands out with lower mean and maximum solving time. 
On the other hand, as illustrated in Table \ref{Comparison of different methods in random dense environments}, compared to the normal scenarios, all the methods in the dense scenarios have lower success rate and longer solving time compared to them in the normal scenarios. Nevertheless, our method still has the highest success rate, shortest mean/max solving time and lowest mean cost of all the benchmarks. To summarize, the proposed PISNO method demonstrates an elevated computational efficiency and capability in trajectory planning.

\begin{figure}[!htbp]
\begin{center}
\subfigure[Planner's performance with different $\alpha$ values.]
{\includegraphics[width=0.48\hsize]{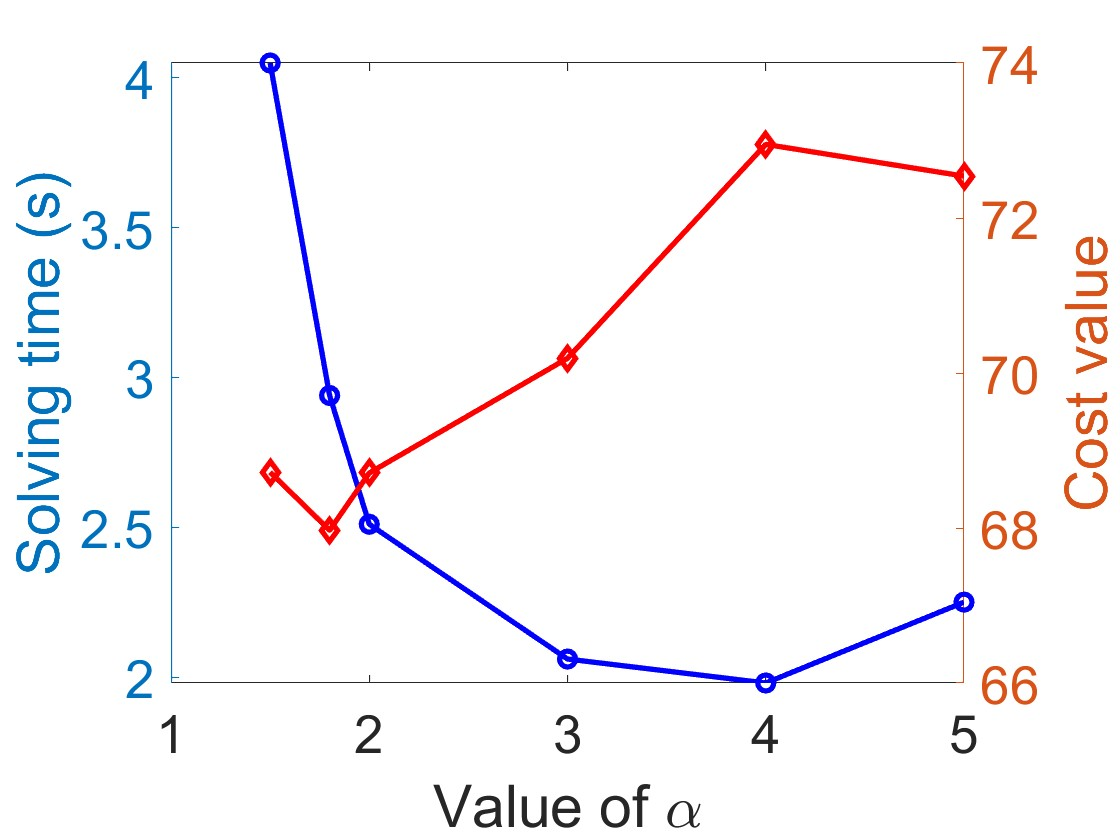}
\label{fig: Planner's performance with different values of alpha.}}
\subfigure[The planning time of PISNO and LIOS + TRMO method in experiments.]
{\includegraphics[width=0.48\hsize]{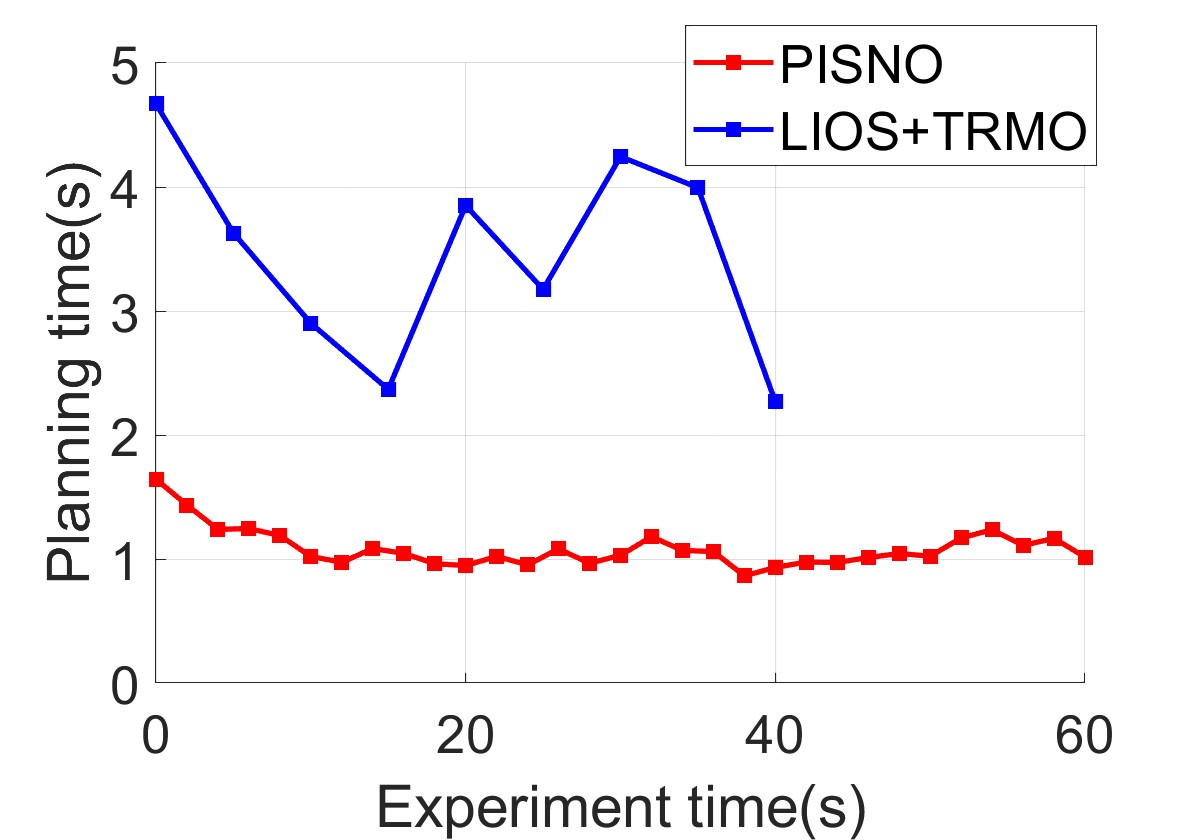}
\label{Planning time in real-world experiments.}}
\caption{Planner's performance with different $\alpha$ values and the planning time in the experiment.}
\label{fig: }
\end{center}
\end{figure}

Moreover, we test the planner's performance with different multiplier $\alpha$ values in simulation, the results are shown in Fig. \ref{fig: Planner's performance with different values of alpha.}. More precisely, as the value of $\alpha$ increases, the difference between the adjacent intermediate OCPs becomes larger, and the planner needs less iteration to reach the threshold sampling number $N_{\rm thre}$. Therefore, our method's solving time would decrease and the cost value would increase.

\begin{figure*}[!htbp]
\centering
\subfigure[PISNO planner, $T=28s$]
{\includegraphics[width=0.42\hsize]{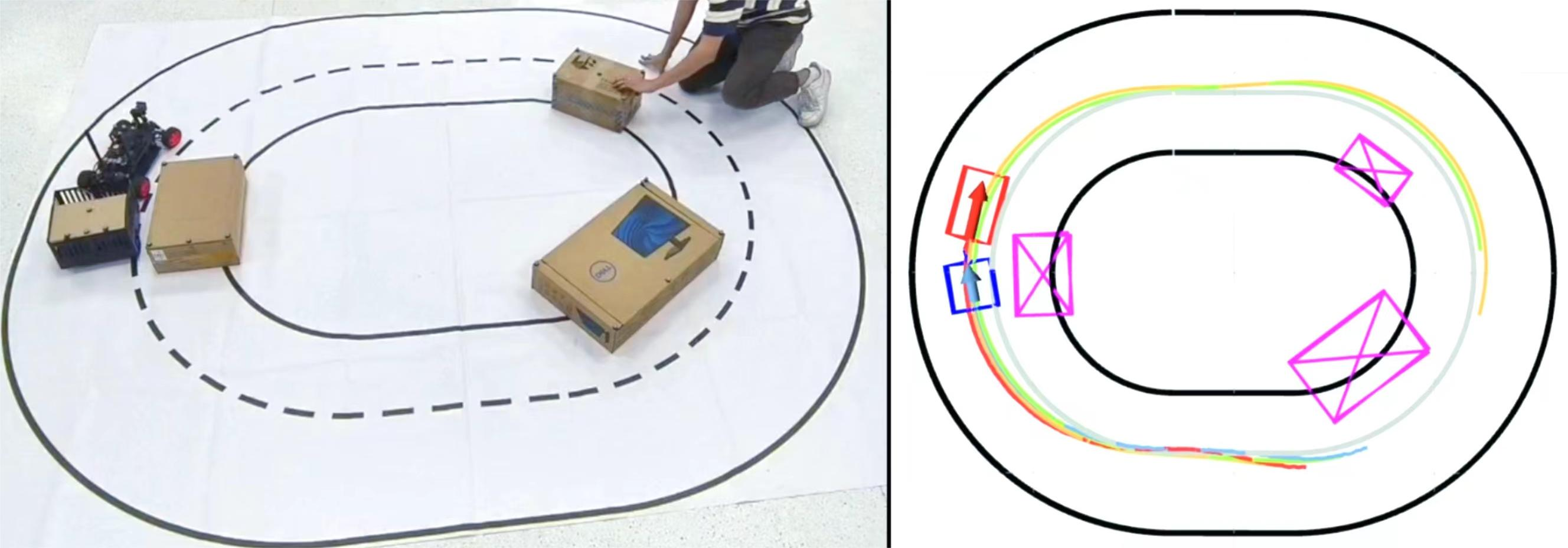}
\label{T=28s}}
\subfigure[PISNO planner, $T=38s$]
{\includegraphics[width=0.42\hsize]{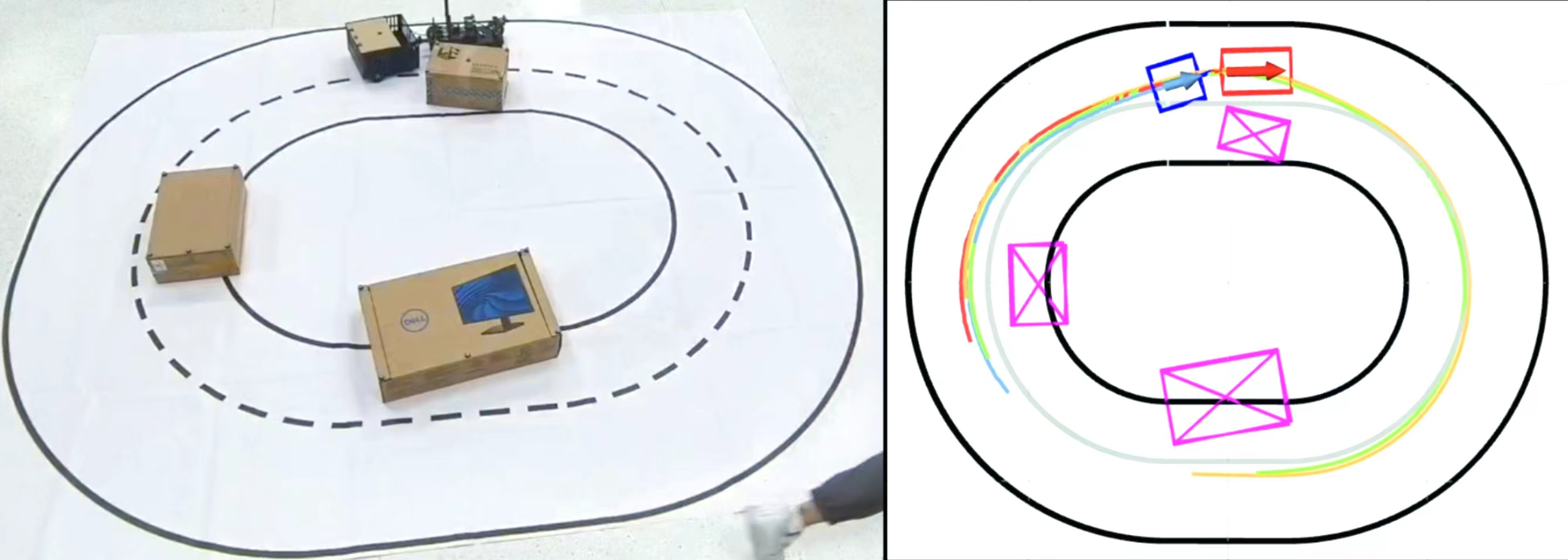}
\label{T=38s}}
\subfigure[LIOS + TRMO planner, $T=28s$]
{\includegraphics[width=0.42\hsize]{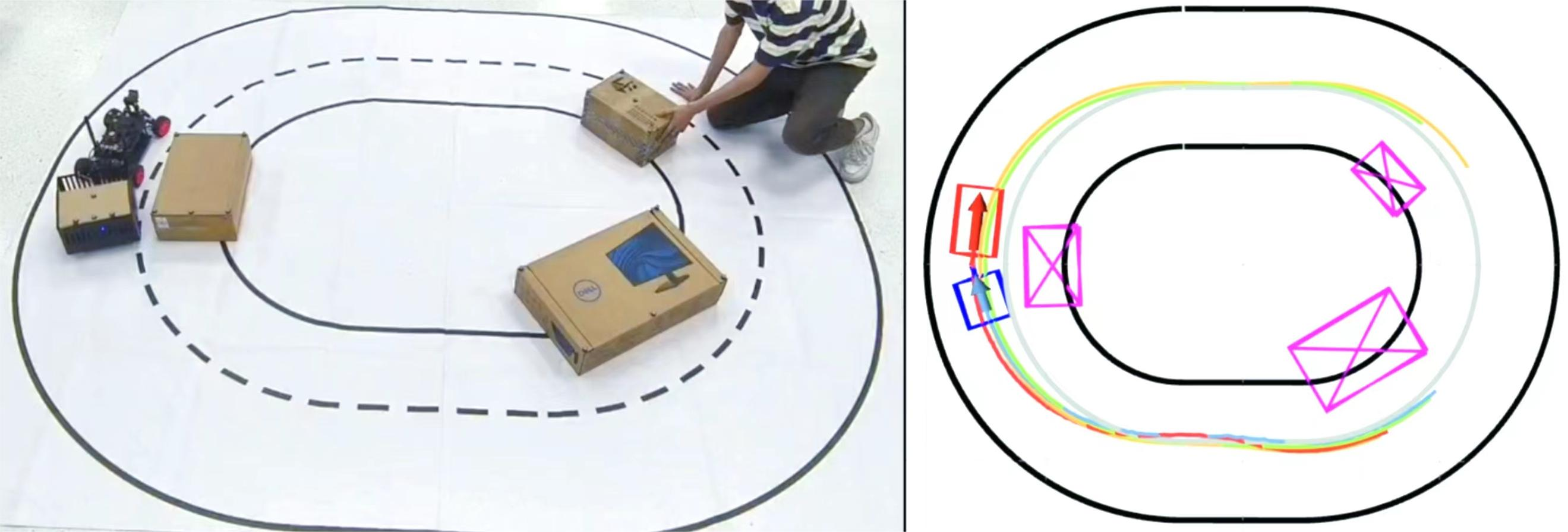}
\label{LIOS + TRMO_T=28s}}
\subfigure[LIOS + TRMO planner, $T=38s$]
{\includegraphics[width=0.42\hsize]{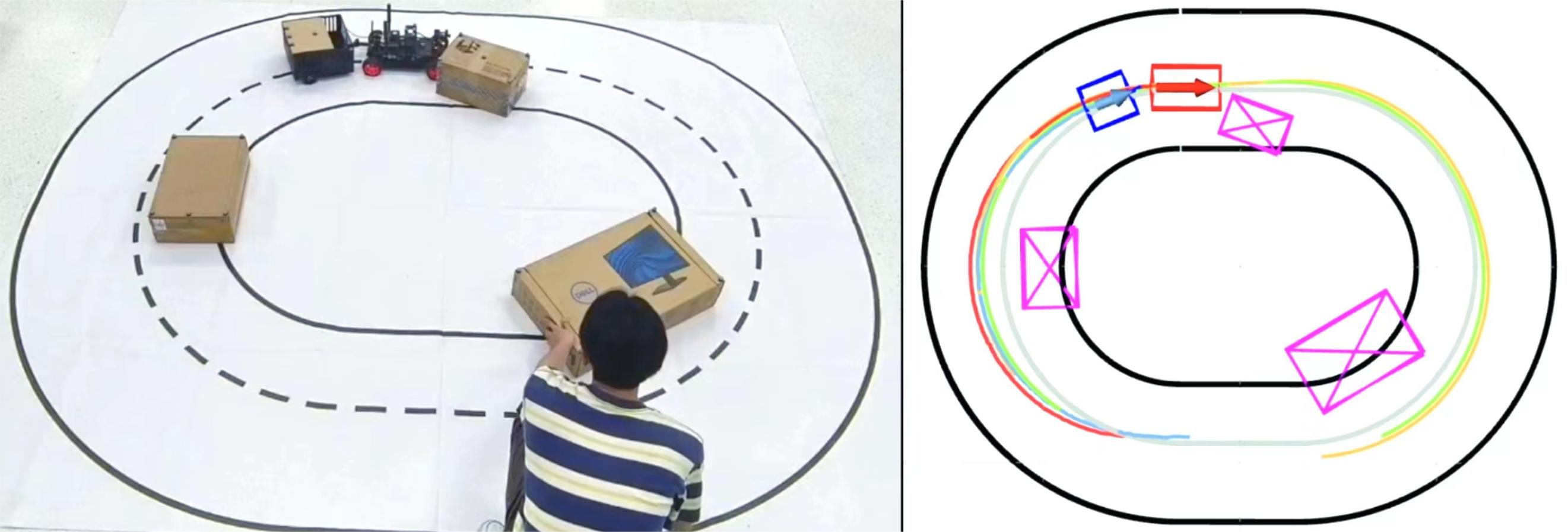}
\label{LIOS + TRMO_T=38s}}
\caption{The results of the experiment at several time-steps of interest, the visualization in rviz are shown in the right part of each pictures. We change the obstacles poses at the time instant $T=28s$ and $T=38s$. PISNO is able to replan the trajectory to avoid collision while LIOS + TRMO fails to do that.}
\label{results of the experiment at several time-steps}
\end{figure*}

\subsection{Experiment results}
\begin{figure}[!htbp]
\centerline{\includegraphics[width=0.7\linewidth]{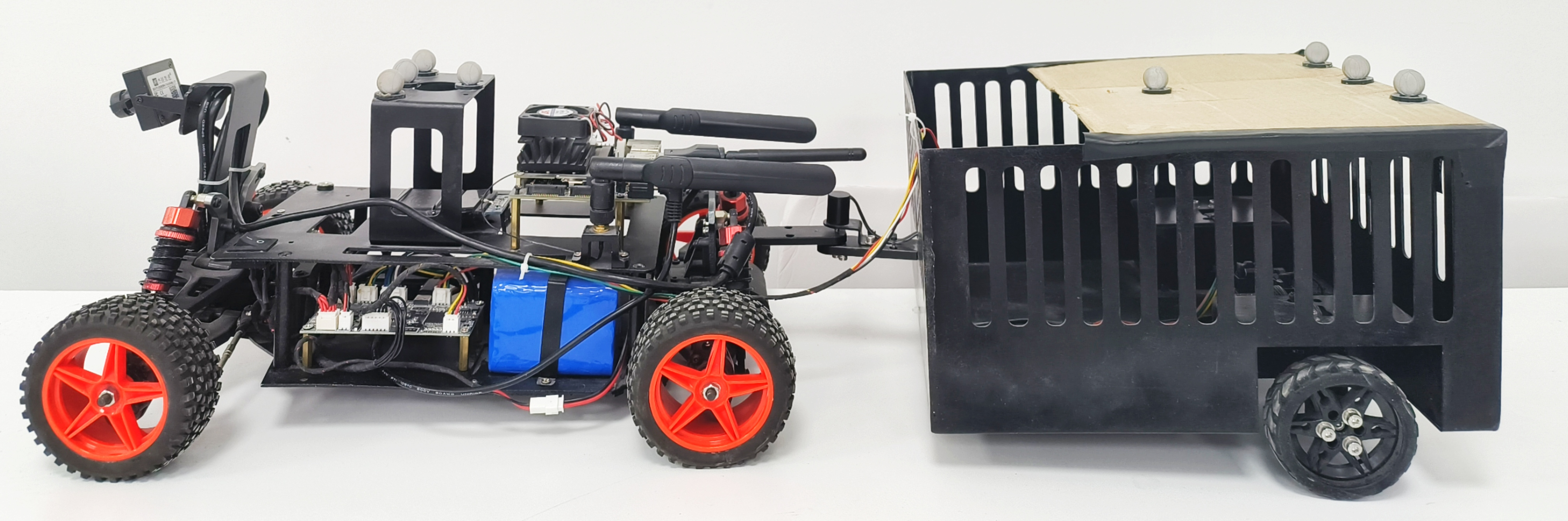}}
\caption{Experiment platform: miniature tractor-trailer model.}
\label{real tractor-trailer}
\end{figure}

Furthermore, the proposed PISNO planner is compared with the LIOS + TRMO planner in real-world. We use a miniature tractor-trailer model as the experiment platform, as shown in Fig. \ref{real tractor-trailer}. 
We use the VICON motion capture system as the localization and perception module to obtain the poses of the tractor-trailer vehicle and the obstacles. The algorithms are run on a laptop with i7-12800HX CPU and 32GB RAM.
In addition, our previous work \cite{K-BMPC} is used to control the vehicle to track the optimized trajectory. 

We compare the performance of the PISNO planner with the LIOS + TRMO planner in a circular scenario, as illustrated in Fig. \ref{results of the experiment at several time-steps}. The poses of the obstacles are changed at the time instants $T=28s$ and $T=38s$. Moreover, we set the replanning time of PISNO to $2s$ and that of LIOS + TRMO to $5s$.
The experiment snap-shots at the time instant $T=28s$ and $T=38s$ are depicted in Fig. \ref{results of the experiment at several time-steps}. At the time instant $T=28s$, as shown in Fig. \ref{T=28s} and \ref{LIOS + TRMO_T=28s}, an obstacle is moved forward to the vehicle. PISNO method adjusts the trajectory according to the movement of the obstacles in time due to its shorter solving time.
As a contrast, LIOS + TRMO method fails to replan a feasible trajectory in time and a collision occurs. 
The planning time for each on-road replanning during the experiment is shown in Fig. \ref{Planning time in real-world experiments.}.
Each planning time of PISNO method is much lower than that of LIOS + TRMO method. Moreover, PISNO's planning time is also lower than the pre-specified replanning time ($2s$). 

\section{Conclusion and future work}

In this work, we present an optimization-based, on-road trajectory planner for tractor-trailer vehicles. 
We first use geometric representations to characterize the lateral and orientation errors in Cartesian frame, and then the errors are used in the cost function and the constraints in the subsequent optimization process. In addition, we generate a coarse trajectory to determine the homotopy class, then we use it as an intuitive initial guess. Furthermore, we propose the PISNO framework that progressively solves the large-scale OCP. Simulation and experiment results demonstrate that the proposed PISNO framework has less solving time over the benchmarks while still maintaining the optimality of the trajectory. Using the previous planned trajectory would further warm-start the optimization process. We will further consider the uncertainties in the trajectory planning and extend our method to the multi-vehicle joint trajectory planning in the future.
\bibliographystyle{IEEEtran}
\bibliography{IEEEabrv,reference}

\begin{thebibliography}{10}
\providecommand{\url}[1]{#1}
\csname url@samestyle\endcsname
\providecommand{\newblock}{\relax}
\providecommand{\bibinfo}[2]{#2}
\providecommand{\BIBentrySTDinterwordspacing}{\spaceskip=0pt\relax}
\providecommand{\BIBentryALTinterwordstretchfactor}{4}
\providecommand{\BIBentryALTinterwordspacing}{\spaceskip=\fontdimen2\font plus
\BIBentryALTinterwordstretchfactor\fontdimen3\font minus \fontdimen4\font\relax}
\providecommand{\BIBforeignlanguage}[2]{{%
\expandafter\ifx\csname l@#1\endcsname\relax
\typeout{** WARNING: IEEEtran.bst: No hyphenation pattern has been}%
\typeout{** loaded for the language `#1'. Using the pattern for}%
\typeout{** the default language instead.}%
\else
\language=\csname l@#1\endcsname
\fi
#2}}
\providecommand{\BIBdecl}{\relax}
\BIBdecl

\bibitem{ritzen2015}
P.~Ritzen, E.~Roebroek, N.~Van De~Wouw, Z.-P. Jiang, and H.~Nijmeijer, ``Trailer steering control of a tractor--trailer robot,'' \emph{IEEE Transactions on Control Systems Technology}, vol.~24, no.~4, pp. 1240--1252, 2015.

\bibitem{zhao2021}
H.~Zhao, S.~Zhou, W.~Chen, Z.~Miao, and Y.-H. Liu, ``Modeling and motion control of industrial tractor--trailers vehicles using force compensation,'' \emph{IEEE/ASME Transactions on Mechatronics}, vol.~26, no.~2, pp. 645--656, 2021.

\bibitem{chai2024cooperative}
R.~Chai, Y.~Guo, Z.~Zuo, K.~Chen, H.-S. Shin, and A.~Tsourdos, ``Cooperative motion planning and control for aerial-ground autonomous systems: Methods and applications,'' \emph{Progress in Aerospace Sciences}, vol. 146, p. 101005, 2024.

\bibitem{wang2020review}
X.~Wang, J.~Liu, X.~Su, H.~Peng, X.~Zhao, and C.~Lu, ``A review on carrier aircraft dispatch path planning and control on deck,'' \emph{Chinese Journal of Aeronautics}, vol.~33, no.~12, pp. 3039--3057, 2020.

\bibitem{li2015robio}
B.~Li and Z.~Shao, ``An incremental strategy for tractor-trailer vehicle global trajectory optimization in the presence of obstacles,'' in \emph{2015 IEEE International Conference on Robotics and Biomimetics (ROBIO)}.\hskip 1em plus 0.5em minus 0.4em\relax IEEE, 2015, pp. 1447--1452.

\bibitem{jujnovich2013}
B.~Jujnovich and D.~Cebon, ``Path-following steering control for articulated vehicles,'' \emph{Journal of Dynamic Systems, Measurement, and Control}, vol. 135, no.~3, p. 031006, 2013.

\bibitem{chai2022design}
R.~Chai, H.~Niu, J.~Carrasco, F.~Arvin, H.~Yin, and B.~Lennox, ``Design and experimental validation of deep reinforcement learning-based fast trajectory planning and control for mobile robot in unknown environment,'' \emph{IEEE Transactions on Neural Networks and Learning Systems}, vol.~35, no.~4, pp. 5778--5792, 2022.

\bibitem{chai2022deep}
R.~Chai, D.~Liu, T.~Liu, A.~Tsourdos, Y.~Xia, and S.~Chai, ``Deep learning-based trajectory planning and control for autonomous ground vehicle parking maneuver,'' \emph{IEEE Transactions on Automation Science and Engineering}, vol.~20, no.~3, pp. 1633--1647, 2022.

\bibitem{chai2020design}
R.~Chai, A.~Tsourdos, A.~Savvaris, S.~Chai, Y.~Xia, and C.~P. Chen, ``Design and implementation of deep neural network-based control for automatic parking maneuver process,'' \emph{IEEE Transactions on Neural Networks and Learning Systems}, vol.~33, no.~4, pp. 1400--1413, 2020.

\bibitem{zhang2024efficient}
R.~Zhang, R.~Chai, K.~Chen, J.~Zhang, S.~Chai, Y.~Xia, and A.~Tsourdos, ``Efficient and near-optimal global path planning for agvs: A dnn-based double closed-loop approach with guarantee mechanism,'' \emph{IEEE Transactions on Industrial Electronics}, 2024.

\bibitem{beyersdorfer2013}
S.~Beyersdorfer and S.~Wagner, ``Novel model based path planning for multi-axle steered heavy load vehicles,'' in \emph{16th International IEEE Conference on Intelligent Transportation Systems (ITSC 2013)}.\hskip 1em plus 0.5em minus 0.4em\relax IEEE, 2013, pp. 424--429.

\bibitem{evestedt2016}
N.~Evestedt, O.~Ljungqvist, and D.~Axehill, ``Motion planning for a reversing general 2-trailer configuration using closed-loop rrt,'' in \emph{2016 IEEE/RSJ International Conference on Intelligent Robots and Systems (IROS)}.\hskip 1em plus 0.5em minus 0.4em\relax IEEE, 2016, pp. 3690--3697.

\bibitem{manav2021}
A.~C. Manav and I.~Lazoglu, ``A novel cascade path planning algorithm for autonomous truck-trailer parking,'' \emph{IEEE Transactions on Intelligent Transportation Systems}, vol.~23, no.~7, pp. 6821--6835, 2021.

\bibitem{zhao2023}
M.~Zhao, T.~Shen, F.~Wang, G.~Yin, Z.~Li, and Y.~Zhang, ``Apten-planner: Autonomous parking of semi-trailer train in extremely narrow environments,'' \emph{IEEE Transactions on Intelligent Transportation Systems}, 2023.

\bibitem{rimmer2016}
A.~J. Rimmer and D.~Cebon, ``Planning collision-free trajectories for reversing multiply-articulated vehicles,'' \emph{IEEE Transactions on Intelligent Transportation Systems}, vol.~17, no.~7, pp. 1998--2007, 2016.

\bibitem{ljungqvist2017}
O.~Ljungqvist, N.~Evestedt, M.~Cirillo, D.~Axehill, and O.~Holmer, ``Lattice-based motion planning for a general 2-trailer system,'' in \emph{2017 IEEE Intelligent Vehicles Symposium (IV)}.\hskip 1em plus 0.5em minus 0.4em\relax IEEE, 2017, pp. 819--824.

\bibitem{wang2023}
Y.~Wang, Z.~Wang, J.~Wang, R.~Guo, and P.~Xiao, ``Pi2-cma based trajectory planning method for the tractor-trailer vehicle,'' in \emph{2023 IEEE 6th International Conference on Pattern Recognition and Artificial Intelligence (PRAI)}.\hskip 1em plus 0.5em minus 0.4em\relax IEEE, 2023, pp. 1230--1235.

\bibitem{leu2022}
J.~Leu, Y.~Wang, M.~Tomizuka, and S.~Di~Cairano, ``Improved a-search guided tree for autonomous trailer planning,'' in \emph{2022 IEEE/RSJ International Conference on Intelligent Robots and Systems (IROS)}.\hskip 1em plus 0.5em minus 0.4em\relax IEEE, 2022, pp. 7190--7196.

\bibitem{li2021}
B.~Li, L.~Li, T.~Acarman, Z.~Shao, and M.~Yue, ``Optimization-based maneuver planning for a tractor-trailer vehicle in a curvy tunnel: A weak reliance on sampling and search,'' \emph{IEEE Robotics and Automation Letters}, vol.~7, no.~2, pp. 706--713, 2021.

\bibitem{li2015}
B.~Li, K.~Wang, and Z.~Shao, ``Time-optimal trajectory planning for tractor-trailer vehicles via simultaneous dynamic optimization,'' in \emph{2015 IEEE/RSJ International Conference on Intelligent Robots and Systems (IROS)}.\hskip 1em plus 0.5em minus 0.4em\relax IEEE, 2015, pp. 3844--3849.

\bibitem{bergman2020}
K.~Bergman, O.~Ljungqvist, and D.~Axehill, ``Improved path planning by tightly combining lattice-based path planning and optimal control,'' \emph{IEEE Transactions on Intelligent Vehicles}, vol.~6, no.~1, pp. 57--66, 2020.

\bibitem{li2019}
B.~Li, T.~Acarman, Y.~Zhang, L.~Zhang, C.~Yaman, and Q.~Kong, ``Tractor-trailer vehicle trajectory planning in narrow environments with a progressively constrained optimal control approach,'' \emph{IEEE Transactions on Intelligent Vehicles}, vol.~5, no.~3, pp. 414--425, 2019.

\bibitem{li2019icra}
B.~Li, Y.~Zhang, T.~Acarma, Q.~Kong, and Y.~Zhang, ``Trajectory planning for a tractor with multiple trailers in extremely narrow environments: A unified approach,'' in \emph{2019 international conference on robotics and automation (ICRA)}.\hskip 1em plus 0.5em minus 0.4em\relax IEEE, 2019, pp. 8557--8562.

\bibitem{cen2021}
H.~Cen, B.~Li, T.~Acarman, Y.~Zhang, Y.~Ouyang, and Y.~Dong, ``Optimization-based maneuver planning for a tractor-trailer vehicle in complex environments using safe travel corridors,'' in \emph{2021 IEEE Intelligent Vehicles Symposium (IV)}.\hskip 1em plus 0.5em minus 0.4em\relax IEEE, 2021, pp. 974--979.

\bibitem{liu2022homogenization}
J.~Liu, X.~Dong, X.~Wang, K.~Cui, X.~Qie, and J.~Jia, ``A homogenization-planning-tracking method to solve cooperative autonomous motion control for heterogeneous carrier dispatch systems,'' \emph{Chinese Journal of Aeronautics}, vol.~35, no.~9, pp. 293--305, 2022.

\bibitem{wang2024safe}
X.~Wang, Z.~Deng, H.~Li, L.~Wang, J.~Jin, and X.~Su, ``Safe dispatch corridor: Towards efficient trajectory planning for carrier aircraft traction system on flight deck,'' \emph{IEEE Transactions on Aerospace and Electronic Systems}, 2024.

\bibitem{li2022}
B.~Li, Y.~Ouyang, L.~Li, and Y.~Zhang, ``Autonomous driving on curvy roads without reliance on frenet frame: A cartesian-based trajectory planning method,'' \emph{IEEE Transactions on Intelligent Transportation Systems}, vol.~23, no.~9, pp. 15\,729--15\,741, 2022.

\bibitem{oliveira2020}
R.~Oliveira, O.~Ljungqvist, P.~F. Lima, and B.~Wahlberg, ``Optimization-based on-road path planning for articulated vehicles,'' \emph{IFAC-PapersOnLine}, vol.~53, no.~2, pp. 15\,572--15\,579, 2020.

\bibitem{shen2021}
Q.~Shen, B.~Wang, and C.~Wang, ``Real-time trajectory planning for on-road autonomous tractor-trailer vehicles,'' \emph{Journal of Shanghai Jiaotong University (Science)}, vol.~26, pp. 722--730, 2021.

\bibitem{oliveira2020iv}
R.~Oliveira, O.~Ljungqvist, P.~F. Lima, and B.~Wahlberg, ``A geometric approach to on-road motion planning for long and multi-body heavy-duty vehicles,'' in \emph{2020 IEEE Intelligent Vehicles Symposium (IV)}.\hskip 1em plus 0.5em minus 0.4em\relax IEEE, 2020, pp. 999--1006.

\bibitem{zhang2020}
Y.~Zhang, H.~Sun, J.~Zhou, J.~Pan, J.~Hu, and J.~Miao, ``Optimal vehicle path planning using quadratic optimization for baidu apollo open platform,'' in \emph{2020 IEEE Intelligent Vehicles Symposium (IV)}.\hskip 1em plus 0.5em minus 0.4em\relax IEEE, 2020, pp. 978--984.

\bibitem{peng2017hp}
H.~Peng, X.~Wang, M.~Li, and B.~Chen, ``An hp symplectic pseudospectral method for nonlinear optimal control,'' \emph{Communications in Nonlinear Science and Numerical Simulation}, vol.~42, pp. 623--644, 2017.

\bibitem{chai2022multiphase}
R.~Chai, A.~Tsourdos, S.~Chai, Y.~Xia, A.~Savvaris, and C.~P. Chen, ``Multiphase overtaking maneuver planning for autonomous ground vehicles via a desensitized trajectory optimization approach,'' \emph{IEEE Transactions on Industrial Informatics}, vol.~19, no.~1, pp. 74--87, 2022.

\bibitem{chai2020multiobjective}
R.~Chai, A.~Tsourdos, A.~Savvaris, S.~Chai, Y.~Xia, and C.~P. Chen, ``Multiobjective overtaking maneuver planning for autonomous ground vehicles,'' \emph{IEEE transactions on cybernetics}, vol.~51, no.~8, pp. 4035--4049, 2020.

\bibitem{chai2020multiobjectiveoptimal}
R.~Chai, A.~Tsourdos, A.~Savvaris, S.~Chai, Y.~Xia, and P.~Chen, ``Multiobjective optimal parking maneuver planning of autonomous wheeled vehicles,'' \emph{IEEE Transactions on Industrial Electronics}, vol.~67, no.~12, pp. 10\,809--10\,821, 2020.

\bibitem{chai2024two}
R.~Chai, K.~Chen, B.~Hua, Y.~Lu, Y.~Xia, X.-M. Sun, G.-P. Liu, and W.~Liang, ``A two phases multiobjective trajectory optimization scheme for multi-ugvs in the sight of the first aid scenario,'' \emph{IEEE Transactions on Cybernetics}, 2024.

\bibitem{lian2023}
J.~Lian, W.~Ren, D.~Yang, L.~Li, and F.~Yu, ``Trajectory planning for autonomous valet parking in narrow environments with enhanced hybrid a* search and nonlinear optimization,'' \emph{IEEE Transactions on Intelligent Vehicles}, 2023.

\bibitem{li2021tits}
B.~Li, T.~Acarman, Y.~Zhang, Y.~Ouyang, C.~Yaman, Q.~Kong, X.~Zhong, and X.~Peng, ``Optimization-based trajectory planning for autonomous parking with irregularly placed obstacles: A lightweight iterative framework,'' \emph{IEEE Transactions on Intelligent Transportation Systems}, vol.~23, no.~8, pp. 11\,970--11\,981, 2021.

\bibitem{IPOPT}
A.~W{\"a}chter and L.~T. Biegler, ``On the implementation of an interior-point filter line-search algorithm for large-scale nonlinear programming,'' \emph{Mathematical programming}, vol. 106, pp. 25--57, 2006.

\bibitem{MA27}
I.~S. Duff and J.~K. Reid, ``The multifrontal solution of indefinite sparse symmetric linear,'' \emph{ACM Transactions on Mathematical Software (TOMS)}, vol.~9, no.~3, pp. 302--325, 1983.

\bibitem{AMPL}
R.~Fourer, D.~M. Gay, and B.~W. Kernighan, ``A modeling language for mathematical programming,'' \emph{Management Science}, vol.~36, no.~5, pp. 519--554, 1990.

\bibitem{K-BMPC}
Z.~Wang, H.~Zhang, and J.~Wang, ``K-bmpc: Derivative-based koopman bilinear model predictive control for tractor-trailer trajectory tracking with unknown parameters,'' \emph{arXiv preprint arXiv:2311.08707}, 2023.

\end{thebibliography}

\end{document}